%% file: icml_paper.tex
\documentclass{article}

\usepackage{microtype}
\usepackage{graphicx}
\usepackage{subfigure}
\usepackage{booktabs} 
\usepackage{multirow}
\usepackage{hyperref}

\usepackage[accepted]{icml2023}

\usepackage{amsmath}
\usepackage{amssymb}
\usepackage{mathtools}
\usepackage{amsthm}

\usepackage[capitalize,noabbrev]{cleveref}

\theoremstyle{plain}

\theoremstyle{definition}

\theoremstyle{remark}

\newcommand{\nzx}[1]{\textcolor{red}{[#1]}}

\usepackage[textsize=tiny]{todonotes}

\icmltitlerunning{Continual Vision-Language Representation Learning with Off-Diagonal Information}

\begin{document}

\twocolumn[
\icmltitle{Continual Vision-Language Representation Learning \\ with Off-Diagonal Information}



\icmlsetsymbol{equla}{*}
\icmlsetsymbol{corre}{\dag}

\begin{icmlauthorlist}
\icmlauthor{Zixuan Ni}{yyy,equla}
\icmlauthor{Longhui Wei}{comp}
\icmlauthor{Siliang Tang}{yyy,corre}
\icmlauthor{Yueting Zhuang}{yyy}
\icmlauthor{Qi Tian}{comp,corre}
\end{icmlauthorlist}

\icmlaffiliation{yyy}{Zhejiang University}
\icmlaffiliation{comp}{Huawei Cloud}

\icmlcorrespondingauthor{Zixuan Ni}{zixuan2i@zju.edu.cn}
\icmlcorrespondingauthor{Siliang Tang}{siliang@zju.edu.cn}
\icmlcorrespondingauthor{Qi Tian}{tian.qi1@huawei.com}

\icmlkeywords{Machine Learning, ICML}

\vskip 0.3in
]



\printAffiliationsAndNotice{\icmlworklocation;\icmlcorresaut} 

\input{content/1_abs.tex}
\input{content/2_intro.tex}

\input{content/3_relate.tex}

\input{content/4_explori.tex}

\input{content/5_reason.tex}
\input{content/6_method.tex}

\input{content/7_experiment_change.tex}

\input{content/8_conclusion.tex}

\section*{Social Impacts}
The goal of continual learning is to help the model adapt to new data domains using only new data, without forgetting its past performance on old data domains. For example, in the fashion field, as fashion trends change, a image-text matching model fitted with old data will gradually become less suitable for the current fashion data. Therefore, by using Mod-X framework, the model can be updated to fit new image-text data using only the current fashion data while preventing the forgetting of past image-text knowledge. Beside of this, catastrophic forgetting, 
arises due to different reasons under various scenarios. In this work, our analysis methods and perspectives on continual image-text pretraining provide new ideas and approaches for future research on different continual learning tasks.

\section*{Acknowledgements}
This work has been supported in part by the Zhejiang NSF (LR21F020004) and the NSFC (No. 62272411). We are grateful to Jiacheng Li and Xin He for their technical assistance. We also appreciate Haizhou Shi and Juncheng Li for their help in writing this paper.




\bibliography{icml_paper}
\bibliographystyle{icml2023}

\newpage
\appendix
\onecolumn
\section{Appendix to Section \ref{explore}}
\label{appendix}

\subsection{The Theoretical Demonstrate that Inter-modal Deviation and Intra-modal Rotation lead to a decline in CLIP’s multimodal retrieval performance}
\label{theoretical}
Inter-modal Deviation and Intra-modal Rotation can influence the CLIP’s sample contrastive matrix, but this does not necessarily lead to errors in multimodal retrieval results. Unless the similarity of the visual language representation of the model for the same sample is smaller than that between different samples. In there, we abstract this problem and give the theoretical conditions that the Intra-modal Rotation and Inter-modal Deviation leads to a performance decline for CLIP on cross-modal retrieval tasks.

There has $N$ image-text pairs \{($\alpha_1$,$\beta_1$),($\alpha_2$,$\beta_2$),($\alpha_3$,$\beta_3$),...,($\alpha_i$,$\beta_i$),...,($\alpha_N$,$\beta_N$)\} $\in$ $R^{W \times W}$. Through function $\mathcal{M}(\alpha)$ and $\mathcal{Q}(\beta)$, $\mathcal{M} \neq \mathcal{Q}$, the Euclidean space A and B of images and texts are formed.

\begin{equation}
\begin{aligned}
& A = span \{\mathcal{M}(\alpha_1),\mathcal{M}(\alpha_2),\mathcal{M}(\alpha_2),...,\mathcal{M}(\alpha_i),...,\mathcal{M}(\alpha_N)\} \\
& B = span \{\mathcal{Q}(\beta_1),\mathcal{Q}(\beta_2),\mathcal{Q}(\beta_2),...,\mathcal{Q}(\beta_i),...,\mathcal{Q}(\beta_N)\}
\end{aligned}
\end{equation}

The $\mathcal{M}(\alpha_i)$,$\mathcal{Q}(\beta_j)$ $\in$ $R^D$ and  $\|\mathcal{M}(\alpha_i)\|=1$ , $\|\mathcal{Q}(\beta_j)\|=1$, $i,j =1,2,3,...,N$. $<\mathcal{M}(\alpha_i),\mathcal{Q}(\beta_j)>$ is the cosine between $\mathcal{M}(\alpha_i)$ and $\mathcal{Q}(\beta_j)$ , $j=1,2,3,...,N$.

\textbf{Suppose}: $\exists (\alpha_a,\beta_a), (\alpha_b,\beta_b)\in 
\{(\alpha_i,\beta_j),i,j = {1,2,3,...,N}\}$ and $a \neq b$ makes:

\begin{equation}
\begin{aligned}
& <\mathcal{M}(\alpha_a),\mathcal{Q}(\beta_a)> ~=~ \mathop{\arg\max}\limits_{\beta_i=\beta_a}<\mathcal{M}(\alpha_a),\mathcal{Q}(\beta_i)> \\
& <\mathcal{M}(\alpha_b),\mathcal{Q}(\beta_b)> ~<~ \mathop{\arg\max}\limits_{\beta_j \neq \beta_b }<\mathcal{M}(\alpha_b),\mathcal{Q}(\beta_j)>
\end{aligned}
\label{inference}
\end{equation}

\subsubsection{How does Inter-modal Deviation affect CLIP's multimodal retrieval performance?}
\label{Inter-modal Deviation}
\textbf{Prove}: There is a rotation matrix pair ($\mathcal{A}$,$\mathcal{B}$) that not only keeps the A and B topology unbiased and makes the 

\begin{equation}
\label{8}
\begin{aligned}
& <\mathcal{M'}(\alpha_a),\mathcal{Q'}(\beta_a)> ~<~ \mathop{\arg\max}\limits_{\beta_i \neq \beta_a}<\mathcal{M'}(\alpha_a),\mathcal{Q'}(\beta_i)> \\
& <\mathcal{M'}(\alpha_b),\mathcal{Q'}(\beta_b)> ~=~ \mathop{\arg\max}\limits_{\beta_j=\beta_b }<\mathcal{M'}(\alpha_b),\mathcal{Q'}(\beta_j)>
\end{aligned}
\end{equation}

where the $\mathcal{M}' = \mathcal{A}(\mathcal{M})$ and $\mathcal{Q}' = \mathcal{B}(\mathcal{Q})$, $\mathcal{A} \neq \mathcal{B}$. And the space A and B can be written as $A'$ and $B'$:

\begin{equation}
\begin{aligned}
& A' = \mathcal{A}(A) = span \{\mathcal{M'}(\alpha_1),\mathcal{M'}(\alpha_2),\mathcal{M'}(\alpha_2),...,\mathcal{M'}(\alpha_i),...,\mathcal{M'}(\alpha_N)\} \\
& B' = \mathcal{B}(B) = span \{\mathcal{Q'}(\beta_1),\mathcal{Q'}(\beta_2),\mathcal{Q'}(\beta_2),...,\mathcal{Q'}(\beta_i),...,\mathcal{Q'}(\beta_N)\}
\end{aligned}
\end{equation}

\textbf{Solution}:
the Equ.\ref{inference} can be written as:

\begin{equation}
\begin{aligned}
& <\mathcal{M}(\alpha_a),\mathcal{Q}(\beta_a)> - <\mathcal{M}(\alpha_a),\mathcal{Q}(\beta_i)> > 0, \forall \beta_i \in \beta, i \neq a\\
& <\mathcal{M}(\alpha_b),\mathcal{Q}(\beta_b)> - <\mathcal{M}(\alpha_b),\mathcal{Q}(\beta_j)> < 0, \exists \beta_j \in \beta, j \neq b
\end{aligned}
\end{equation}

hence:
\begin{equation}
\label{11}
\begin{aligned}
& \mathcal{M}(\alpha_a)^T\mathcal{Q}(\beta_a) - \mathcal{M}(\alpha_a)^T\mathcal{Q}(\beta_i) > 0, \forall \beta_i \in \beta, i \neq a\\
& \mathcal{M}(\alpha_b)^T\mathcal{Q}(\beta_j) - \mathcal{M}(\alpha_b)^T\mathcal{Q}(\beta_b) > 0, \exists \beta_j \in \beta, j \neq b
\end{aligned}
\end{equation}

because the rotation matrix pair ($\mathcal{A}$,$\mathcal{B}$) can be seen as a rotation matrix $\mathcal{R}(\theta^D)$, where the $\theta^D$ is a rotation angle between AB and $A'B'$. Hence, when applying this rotation matrix $\mathcal{R}(\theta^D)$, the Equ.\ref{11} can be written as:

\begin{equation}
\begin{aligned}
& \mathcal{M}(\alpha_a)^T\mathcal{R}(\theta^D)\mathcal{Q}(\beta_a) - \mathcal{M}(\alpha_a)^T\mathcal{R}(\theta^D)\mathcal{Q}(\beta_i) < 0, \exists \beta_i \in \beta, i \neq a\\
& \mathcal{M}(\alpha_b)^T\mathcal{R}(\theta^D)\mathcal{Q}(\beta_j) - \mathcal{M}(\alpha_b)^T\mathcal{R}(\theta^D)\mathcal{Q}(\beta_b) < 0, \forall \beta_j \in \beta, j \neq b
\end{aligned}
\label{12}
\end{equation}

Because the rotation matrix satisfies that the inner product of itself is 1. So, Equ \ref{12} can be written as:

\begin{equation}
\mathcal{M}(\alpha_a)^T\mathcal{R}(\theta^D)(\mathcal{Q}(\beta_a) - \mathcal{Q}(\beta_i)) < 0, \exists \beta_i \in \beta, i \neq a,  \mathcal{R}(\theta^D)^T\mathcal{R}(\theta^D) = \mathcal{I}
\label{13}
\end{equation}
\begin{equation}
\mathcal{M}(\alpha_b)^T\mathcal{R}(\theta^D)(\mathcal{Q}(\beta_j) - \mathcal{Q}(\beta_b)) < 0, \forall \beta_j \in \beta, j \neq b,  \mathcal{R}(\theta^D)^T\mathcal{R}(\theta^D) = \mathcal{I}
\label{14}
\end{equation}

For example, when $\mathcal{R}(\theta^D) = -\mathcal{I}$, then $\mathcal{R}(\theta^D)^T\mathcal{M}(\alpha_a) = -\mathcal{M}(\alpha_a)$ the equ \ref{13} and \ref{14} will hold. So, rotation matrices ($\mathcal{A}$,$\mathcal{B}$) that makes Equ.\ref{8} true exists.

\subsubsection{How does Intra-modal Rotation affect CLIP's multimodal retrieval performance?}
\label{Intra-modal Rotation}
Since intra-modal rotation just requires the length of representation vectors after rotation is 1 and does not require that the intra-modal representation space is invariant, it is a more general case of inter-modal deviation. This means that all rotation matrixes that satisfy \ref{Inter-modal Deviation} can also satisfy Intra-modal Rotation. Different from intra-modal deviation, the inner product of the mapping matrix $\mathcal{P}$ does not require to be 1. So, we rewrite the Equ \ref{13} and \ref{14} to:
\begin{equation}
\mathcal{M}(\alpha_a)^T(\mathcal{Q}(\beta_a) - \mathcal{Q}(\beta_i))\mathcal{P} < 0, \exists \beta_i \in \beta, i \neq a,
\label{13}
\end{equation}
\begin{equation}
\mathcal{M}(\alpha_b)^T(\mathcal{Q}(\beta_j) - \mathcal{Q}(\beta_b))\mathcal{P} < 0, \forall \beta_j \in \beta, j \neq b,
\label{14}
\end{equation}

Any mapping matrix $\mathcal{P}$ that rotation the direction of $(\mathcal{Q}(\beta_j) - \mathcal{Q}(\beta_b))$ by more than 90 degrees.

\subsection{Detailed SAM and RAM Distribution of Language Encoders}
\label{detail_language_encoder}
The topology of the language representation space does not change significantly during the continual CLIP training. But the whole language representation space, like the vision representation space, has a large rotation around the center of the high-dimensional sphere during the continual training. The angle change distribution Table \ref{table:text_analysis}(a) and rotation angle distribution Table \ref{table:text_analysis}(b) are shown below. 

\begin{figure}[!h]
\small
\begin{center}
\subfigure[The include angle distribution (SAM) in text representation space.]{
\begin{minipage}[c]{\linewidth}
\center
\small
\begin{tabular}{|c|c|c|c|c|c|}
\toprule
 $\theta_{SAM} \in$ & $[0^{\circ}$, $5^{\circ}]$ & $(5^{\circ}$, $10^{\circ}]$ & $(10^{\circ}$, $15^{\circ}]$ & $(15^{\circ}$, $20^{\circ}] $ & $(20^{\circ}$, $180^{\circ}]$ \\
\midrule
$SAM_{0 - 1}$ &  64.43$\%$ & 28.49$\%$& 6.23$\%$ & 0.78$\%$ & 0.07$\%$ \\
$SAM_{1 - 2}$ &  71.54$\%$ & 24.89$\%$& 3.35$\%$ & 0.22$\%$ & 0.01$\%$ \\
$SAM_{2 - 3}$ &  71.36$\%$ & 25.01$\%$& 3.40$\%$ & 0.22$\%$ & 0.01$\%$ \\
$SAM_{3 - 4}$ &  67.30$\%$ & 27.27$\%$& 4.93$\%$ & 0.48$\%$ & 0.03$\%$ \\
$SAM_{4 - 5}$ &  58.84$\%$ & 30.70$\%$& 8.77$\%$ & 1.50$\%$ & 0.20$\%$ \\
$SAM_{0 - 5}$ &  55.39$\%$ & 31.60$\%$& 10.52$\%$ & 2.15$\%$ & 0.33$\%$ \\
\bottomrule
\end{tabular}
\end{minipage}
\label{table:text_analysis_a}
}

\subfigure[The rotation angle distribution (RAM) in text representation space.]{
\begin{minipage}[c]{\linewidth}
\center
\small
\begin{tabular}{|c|c|c|c|c|c|}
\toprule
 $\theta_{RAM} \in$ & $[0^{\circ}$, $15^{\circ}]$ & $(15^{\circ}$, $20^{\circ}]$ & $(20^{\circ}$, $25^{\circ}]$ & $(25^{\circ}$, $30^{\circ}] $ & $(30^{\circ}$, $180^{\circ}]$ \\
\midrule
$RAM_{(0,1)}$ &  0.00$\%$ & 1.94$\%$& 28.38$\%$ & 45.88$\%$ & 23.80$\%$ \\
$RAM_{(1,2)}$ &  0.02$\%$ & 8.90$\%$& 47.76$\%$ & 34.94$\%$ & 8.38$\%$ \\
$RAM_{(2,3)}$ &  0.04$\%$ & 1.14$\%$& 49.86$\%$ & 31.18$\%$ & 7.52$\%$ \\
$RAM_{(3,4)}$ &  0.02$\%$ & 2.84$\%$& 33.70$\%$ & 43.76$\%$ & 19.68$\%$ \\
$RAM_{(4,5)}$ &  0.00$\%$ & 0.04$\%$& 3.28$\%$ & 27.66$\%$ & 69.02$\%$ \\
$RAM_{(0,5)}$ &  0.00$\%$ & 0.00$\%$& 0.00$\%$ & 1.12$\%$ & 98.88$\%$ \\
\bottomrule
\end{tabular}
\end{minipage}
\label{table:text_analysis_b}
}
\caption{Detailed SAM and RAM Distribution of Language Encoders.}
\label{table:text_analysis}
\end{center}
\end{figure}

By observing the table in Table \ref{table:text_analysis}, we can find that more than 88\% of the angle change between any two language representation vectors in the language representation space are between 0 and 10 degrees in the process of continual CLIP training, while only 20\% are above 10 degrees. Moreover, less than 0.2\% of the angle changes is above 20 degrees. Those angle change between 15-20 degrees also only account for about 1.5\% of all images pairs. Similar to the visual representation space, the direction of the same sample in the language representation space of different training phases also has changed greatly. However, unlike most of the rotations in the vision representation space, which are distributed over 30 degrees, in the language space, the rotations in the representation space are mostly distributed between 20 and 30 degrees. \textbf{Because of this difference, the representation alignment of the CLIP for different modalities of the same sample deviates during the continual training.}

\subsection{The representation quality of vision encoders during continual CLIP training}
\label{representation_quality}
\begin{figure}[!h]
    \centering
	\centering
	\includegraphics[width=3in]{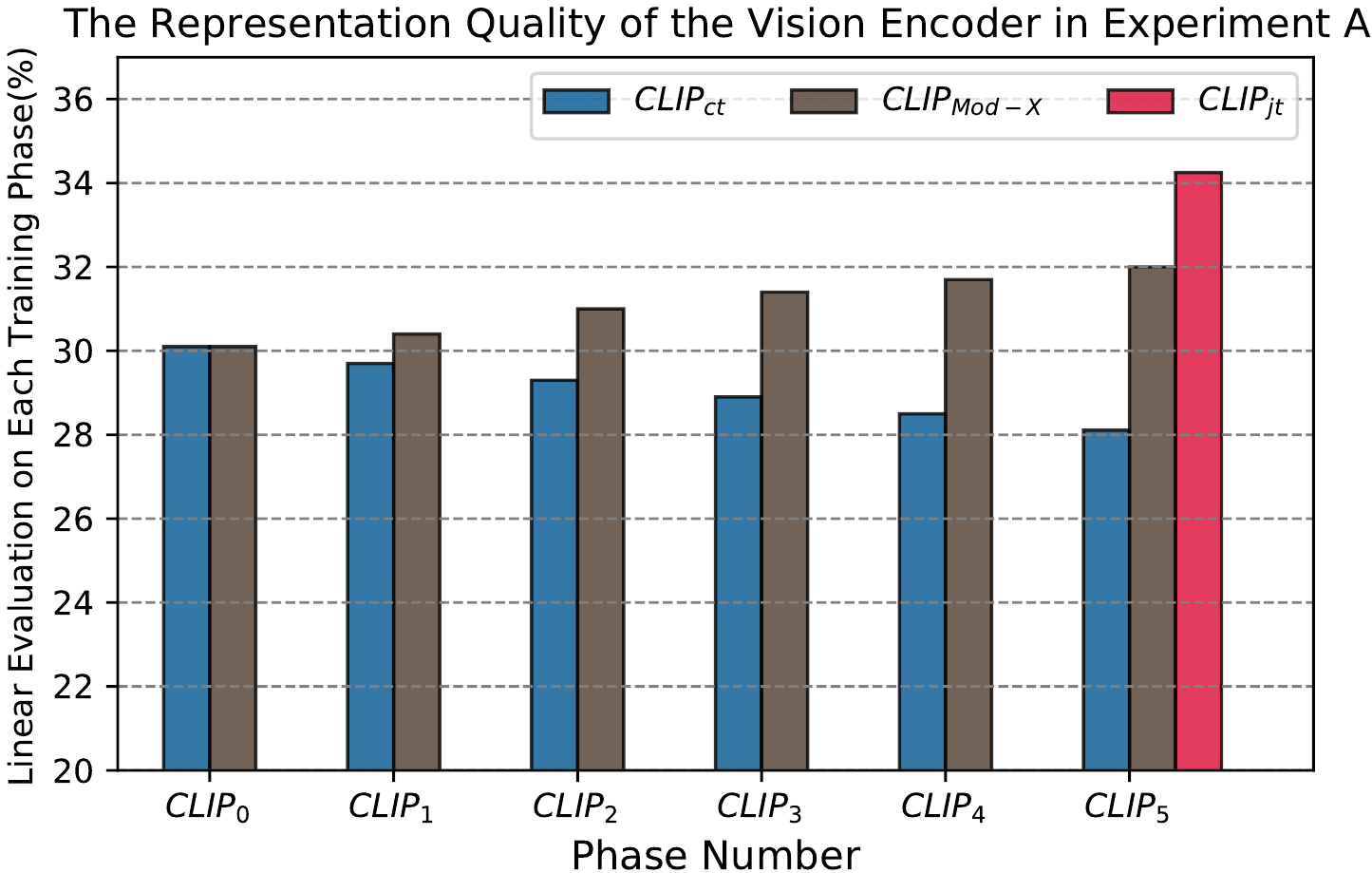}
	\caption{The representation quality of visual encoders in each training phase in explore experiment \ref{explore}.}
	\label{A_linear}
\end{figure}

In Section \ref{explore}, based on the distribution Table \ref{SAM_b}, we inference that the topology of the visual representation of the CLIP$_{ct}$ changes slowly during the continual CLIP training. Due to the topology of the representation space is correlated with the quality of the model's representation, so we use the linear probe evaluation method, commonly used in self-supervision \cite{oord2018representation,he2020momentum}, to detect the quality of the model's vision encoders to verify our suppose.By fixing the vision encoder, retrain a single Linear layer, which is connected behind the vision encoder, based on the ImageNet \cite{deng2009imagenet} training set and evaluate its top-1 accuracy on the ImageNet test set to represent the vision encoder's representation quality. As shown in Figure \ref{A_linear}, we calculate the vision encoders' linear evaluation in each training phase in explore experiment \ref{explore}. 

Observing the changing trends in the linear evaluation accuracy of each training phase, we can find that the representation quality of the vision encoder in CLIP$_{cl}$ gradually decreases as the training phase increases. The top-1 accuracy in the ImageNet test set dropped from 30.1\% to 28.1\%, which is consistent with our conjecture \ref{ImR}. Compared to the decline in multimodal retrieval, the decrease in the quality of visual representations appears to be negligible. In addition, by comparing the results of CLIP$_{Mod-X}$ and CLIP$_{jt}$, we can find that our Mod-X framework can not only help the model fit new image-text samples but also improve the representation quality of the modal encoders within the CLIP. The top-1 accuracy of the vision encoder in CLIP$_{Mod-X}$ improved from 30.1\% to 32.0\%. All of this also illustrates that the quality of the extractor representation is not precisely positively correlated with the decline in multimodal retrieval performance of CLIP model. Alignment of the representation between the different modals is also critical.

\section{Appendix to Section \ref{experiments}}
\label{Appendix_to_experiments}
\subsection{Detailed Experiment Setting}
\label{Experiment_Setting}
In exploration experiments \ref{explore} and Experiment \ref{experiment_A}, we use RN50 \cite{he2016deep} as the vision encoder. In experiment \ref{Experiments_ECommerce-T2I} we use Vit-32/B as the vision encoder. The language encoder in all experiments is a transformer-based architecture which follows modification proposed in CLIP \cite{OpenAi}. In all experiments, the input images are resized to 224 × 224 and the input texts are tokenized by WordPiece with a maximum length of 77. We utilize AdamW \cite{loshchilov2017decoupled} optimizer and a cosine annealing learning rate schedule with warmup which is consistent with \cite{OpenAi}. All of the experiments are conducted on 8 NVIDIA V100 GPUS. 

In exploration experiment \ref{explore} and Experiment \ref{experiment_A}, we use the hyper-parameters as be shown in table \ref{hyparams_A}. Since the experiment \ref{Experiments_ECommerce-T2I} based on the pre-training model ViT-32/B in \cite{OpenAi}, we set a smaller learning rate from 5e-4 to 1e-6. And other hyper-parameters is consistent with Experiment \ref{experiment_A} and CLIP \cite{OpenAi}.
\begin{table*}[h!]
    \centering
    \subfigure[]{
    \begin{minipage}[]{0.48\linewidth}
    \center
    \begin{tabular}{lc}
    \toprule
    \textbf{Hyperparameter}            & \textbf{Value}     \\
    \midrule
    Batch size                         & 280               \\
    Vocabulary size                    & 49408              \\
    Training epochs                    & 35                \\
    Initial temperature $\tau$         & 0.07               \\
    $\alpha$                           & 20                 \\
    Weight decay                       & 0.2              \\
    Warm-up iterations (\%)            & 20                \\
    Learning rate                      & $5e^{-4}$ \\
    Adam $\beta_1$                     & 0.9                \\
    Adam $\beta_2$                     & 0.99               \\
    Adam $\epsilon$                    & $1e^{-8}$          \\
    \bottomrule
    \end{tabular}
    \end{minipage}
    \label{hyparams_A}
    }
    \subfigure[]{
    \begin{minipage}[c]{0.48\linewidth}
    \center
    \begin{tabular}{lc}
    \toprule
    \textbf{Hyperparameter}            & \textbf{Value}     \\
    \midrule
    Batch size                         & 280               \\
    Vocabulary size                    & 49408              \\
    Training epochs                    & 35                \\
    Initial temperature $\tau$         & 0.07               \\
    $\alpha$                           & 20                 \\
    Weight decay                       & 0.2              \\
    Warm-up iterations (\%)            & 20                \\
    Learning rate                      & $1e^{-6}$ \\
    Adam $\beta_1$                     & 0.9                \\
    Adam $\beta_2$                     & 0.99               \\
    Adam $\epsilon$                    & $1e^{-8}$          \\
    \bottomrule
    \end{tabular}
    \end{minipage}
    \label{hyparams_B}
    }
    \caption{Table (a) is the hyper-parameter in exploration experiment (Section \ref{explore}) and Experiment \ref{experiment_A}. Table (b) is the hyper-parameter in Experiment \ref{Experiments_ECommerce-T2I}).}
\end{table*}


\subsection{The Relationship Between Contrastive Matrix, Intra-modal Rotation, Inter-modal Deviation and Mod-X}
\label{relationship_cm_ims_modx}
From a detailed point of view, the element $M_{i,j}$ in the $i$,$j$ position of the contrastive matrix $M$ is the similarity score of the $i$'th sample vision embedding and the $j$'th sample text embedding. Since the length of the representation vector is \textbf{1}, the similarity score $M_{i,j}$ also refers to the angle between the $i$'th sample vision embedding and the $j$'th sample text embedding. Greater similarity means a smaller angle. Therefore, the value of the diagonal elements in the contrast matrix $M$ represents the angle between different modals of the same sample. The value of the off-diagonal elements represents the angle between the different modals of different samples in the CLIP's representation space. Through our exploration (in section \ref{explore}), the Intra-modal Rotation and the Inter-modal Deviation affect these angles or similarity scores. From an overall perspective, \textbf{the similarity distribution of the contrastive matrix $M$ is equivalent to the structure of the representation space of the model.} Our Mod-X framework attempts to distill the similarity distribution of off-diagonal elements identical to aligning the model's representation space structure, which reduces the influence of spatial disorder during continual CLIP training.

\begin{figure}[!h]
\vspace{-0.5em}
\subfigure[]{
\begin{minipage}{0.4\linewidth}
\centering
\includegraphics[width = 0.9\linewidth]{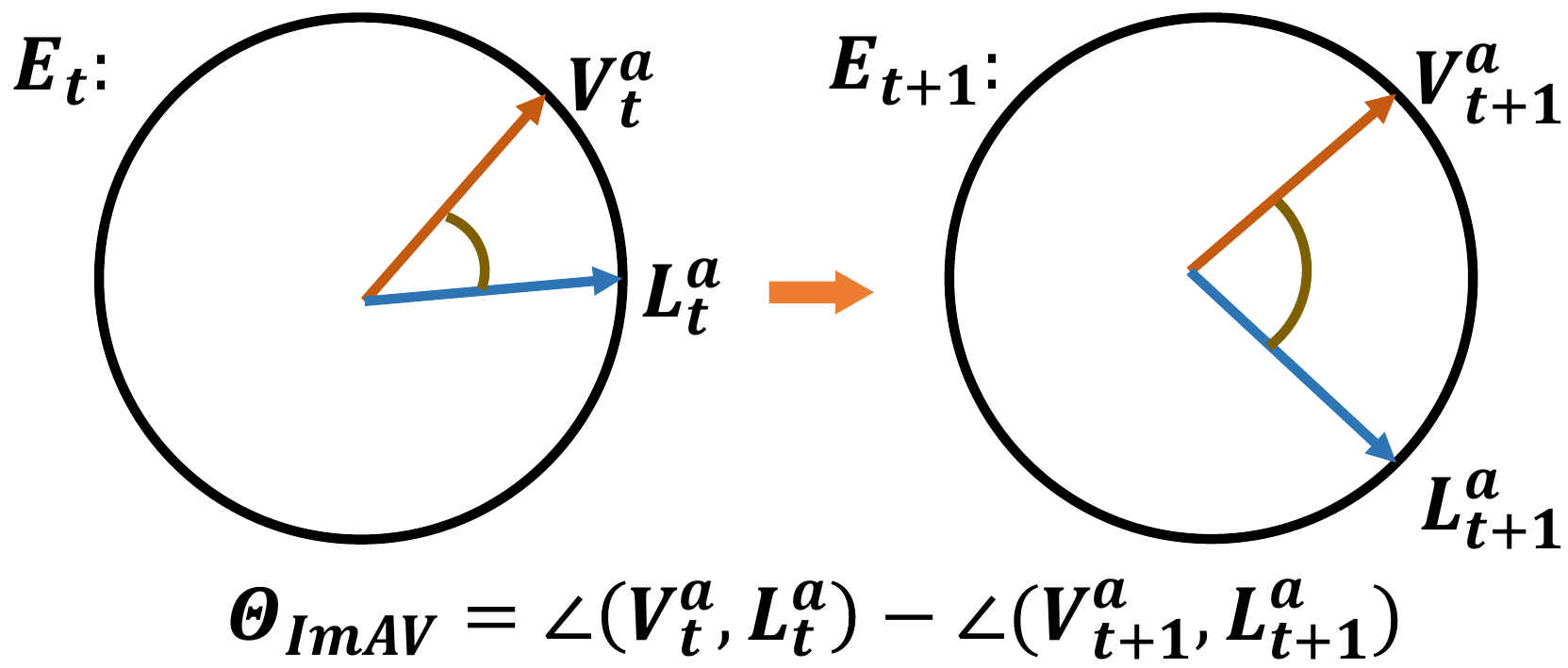}
\label{ImAM_a}
\end{minipage}
}
\subfigure[]{
\setlength\tabcolsep{2pt}
\begin{minipage}[r]{0.6\linewidth}
\footnotesize
\begin{tabular}{|c|c|c|c|c|c|}
\toprule
 $\theta_{ImAV} \in$ & $[0^{\circ}$, $5^{\circ}]$ & $(5^{\circ}$, $10^{\circ}]$ & $(10^{\circ}$, $15^{\circ}]$ & $(15^{\circ}$, $20^{\circ}] $ & $(20^{\circ}$, $180^{\circ}]$ \\
\midrule
$ImAV_{(0,1)}$ &  44.78$\%$ & 31.54$\%$& 14.62$\%$ & 8.26$\%$ & 0.81$\%$ \\
$ImAV_{(1,2)}$ &  50.37$\%$ & 28.48$\%$& 16.58$\%$ & 4.57$\%$ & 0.00$\%$ \\
$ImAV_{(2,3)}$ &  49.70$\%$ & 24.22$\%$& 20.53$\%$ & 5.13$\%$ & 0.42$\%$ \\
$ImAV_{(3,4)}$ &  46.25$\%$ & 30.12$\%$& 18.53$\%$ & 4.81$\%$ & 0.29$\%$ \\
$ImAV_{(4,5)}$ &  43.36$\%$ & 32.83$\%$& 19.81$\%$ & 3.82$\%$ & 0.18$\%$ \\
$ImAV_{(0,5)}$ &  31.98$\%$ & 33.62$\%$& 24.37$\%$ & 10.01$\%$ & 0.02$\%$ \\
\bottomrule
\end{tabular}
\label{ImAM_b}
\end{minipage}
}
\vspace{-1em}
\caption{The sub-figure on the left shows a schematic diagram of computing $\theta_{ImAV}$. The table on the right shows the distribution of the included angle change between the vision and language representation of the samples in CLIP$_{ct}$, which were correctly retrieved in the previous training phases.}
\label{ImAV}
\vspace{-0.5em}
\end{figure}

To better illustrate the relationship between the model's representation space and the model's similarity performance, we add a more direct statistical analysis, \textbf{inter-modal angle variation distribution}. Based on the settings in section \ref{explore}, in the training phase $t$, we compare the change of angle distribution between modalities for the training samples retrieved correctly in the training phase $t-1$. A schematic diagram of inter-modal angle variation $\theta_{ImAV}$ is shown in Figure \ref{ImAM_a}, where the sample $a$ refers to the training sample that can be retrieved correctly by model CLIP$_{t-1}$ in training phase $t-1$. The $V$ is the vision representation and $L$ is the language representation. Inter-modal angle variation distribution table can be seen in Figure \ref{ImAM_b}.

As shown in Figure \ref{ImAM_b}, during the continual training, the samples that were correctly retrieved in the past have apparent changes in the angle between the modalities as the training phases go up. Only less than 50\% of the samples change within 5 degrees in the continual training, and about 30\% of the samples have a change of 5-10 degrees. However, more than 20\% of the samples change their included angle by more than 10 degrees during the training process. This shows that the inter-modal spatial alignment (similarity performance) of the CLIP$_{ct}$ is affected by spatial disorder. 

To illustrate our Mod-X framework indeed alleviates the spatial disorder between sample's modalities during continual training, we show the inter-modal angle variation distribution of the CLIP$_{Mod-X}$ in Experiment \ref{experiment_A} in Table \ref{ImAV_modx}.
\begin{table*}[!h]
\small
\begin{center}
\setlength\tabcolsep{1.8pt}
\begin{tabular}{|c|c|c|c|c|c|}
\toprule
 $\theta_{ImAM} \in$ & $[0^{\circ}$, $5^{\circ}]$ & $(5^{\circ}$, $10^{\circ}]$ & $(10^{\circ}$, $15^{\circ}]$ & $(15^{\circ}$, $20^{\circ}] $ & $(20^{\circ}$, $180^{\circ}]$ \\
\midrule
$ImAM_{(0,1)}$ &  88.66$\%$ & 7.81$\%$& 2.56$\%$ & 0.97$\%$ & 0.00$\%$ \\
$ImAM_{(1,2)}$ &  91.79$\%$ & 4.01$\%$& 3.20$\%$ & 0.00$\%$ & 0.00$\%$ \\
$ImAM_{(2,3)}$ &  90.70$\%$ & 9.02$\%$& 0.24$\%$ & 0.04$\%$ & 0.01$\%$ \\
$ImAM_{(3,4)}$ &  92.13$\%$ & 6.20$\%$& 1.61$\%$ & 0.06$\%$ & 0.00$\%$ \\
$ImAM_{(4,5)}$ &  91.91$\%$ & 7.71$\%$& 0.38$\%$ & 0.00$\%$ & 0.00$\%$ \\
$ImAM_{(0,5)}$ &  87.81$\%$ & 10.87$\%$& 1.12$\%$ & 0.20$\%$ & 0.00$\%$ \\
\bottomrule
\end{tabular}
\caption{The table shows the distribution of the included angle change between the vision and language representation of the sample in the CLIP$_{Mod-X}$ in Experiment \ref{experiment_A}.}
\label{ImAV_modx}
\end{center}
\end{table*}

Comparing the Figure \ref{ImAM_b} and Table \ref{ImAV_modx}, it can be found that the CLIP$_{Mod-X}$ well maintains the inter-modal spatial alignment of the correctly retrieved samples during the continual CLIP training. On average, 90\% of the correctly retrieved samples have an angle change of less than 5 degrees in continual training, and the samples with an angle change of more than 15 degrees account for less than 1\% of all samples. All of this shows that the Mod-X framework does mitigates the spatial disorder during continual CLIP training by preserving the inter-modal spatial alignment of the samples retrieved correctly in the past during the continual training.

\subsection{Validation of Inter-modal Deviation on ECommerce-T2I dataset}
\label{val_on_EC}
\begin{figure}[h]
	\centering
	\includegraphics[height=2.5in]{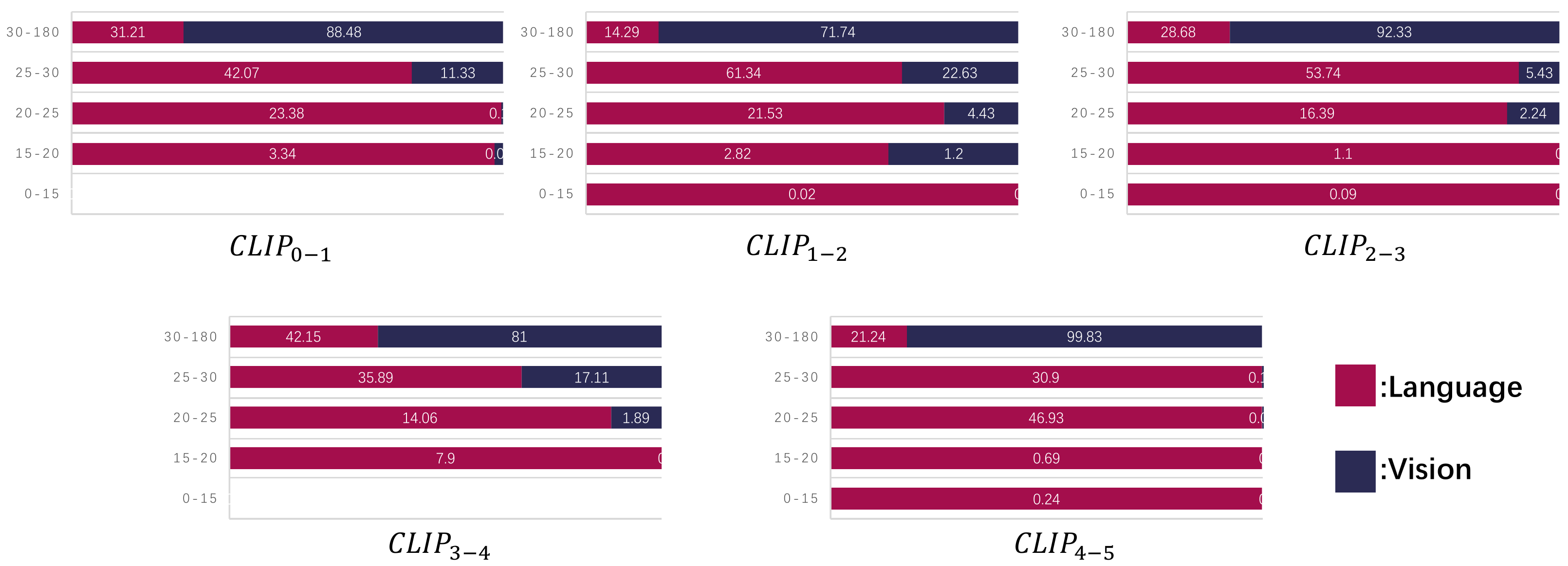}
	\caption{The comparison of the rotation distributions of the vision encoder and langugae encoder during continual CLIP training on ECommerce-T2I dataset. CLIP$_{i-j}$ refers to the CLIP's continual training from training phase $i$ to $j$. The values under the same color represent the proportion of test samples to total samples in each rotation angle interval of the same modality.}
	\vspace{-0.5em}
	\label{difference_RAM_EC}
\end{figure}

In section \ref{explore}, we discuss the representational space variation of CLIP$_{ct}$ under the open-world dataset COCO\cite{lin2014microsoft} and Flickr30K\cite{young2014image}. In there, following the explore settings of the section \ref{ImD}, we compare the rotation distribution of the representation space of the vision and language extractors of CLIP$_{ct}$ under the specific e-commerce text to image dataset ECommerce-T2I \cite{M6-T} (Experiment \ref{Experiments_ECommerce-T2I})  By evaluating the rotation distribution of the modal's representation space at various training phases on the COCO(5K) testset, we drawn the rotation distribution comparison diagram in Figure \ref{difference_RAM_EC}.

From Figure \ref{difference_RAM_EC}, we can find that when the CLIP is trained on a specific data domain, the rotation of visual representation space becomes more severe, among which more than 70\% of the samples have more than 30 degrees of rotation in the visual space, which is higher than that of the open-world dataset. Although the rotation of more than 30 degrees in the language space has also seen a large proportional increase than the open-world dataset, it is still significantly out of sync with the rotation in the visual space. Most samples are rotated within 30 degrees in language space. Through this validation, we show that inter-modal deviation (rotational asynchrony) of the representation space of different modal encoders persists during the continual CLIP training on a specific data domain.

\subsection{The sensitivity of hyper-parameter $\alpha$}
\label{sensitivity_hp}
In this section, we discuss the effect of different $\alpha$ on the final performance of the CLIP$_{Mod-X}$ based on the settings of Experiment \ref{experiment_A}. Table \ref{alpha_modx} presents the final retrieval results of the CLIP$_{Mod-X}$ model with $\alpha$ $=$ 10, 15, 20, 25, 30.

\begin{table*}[!h]
\small
\begin{center}
\setlength\tabcolsep{1.9pt}
\begin{tabular}{c|l|ccc|ccc|ccc|ccc}
\toprule
\multirow{3}{*}{\shortstack[c]{Pretraining\\ Dataset}} & \multirow{3}{*}{\shortstack[c]{Model}} & \multicolumn{6}{c}{Image-Text Retrieval(\%)} & \multicolumn{6}{c}{Text-Image Retrieval(\%)} \\
& & \multicolumn{3}{c}{Flickr30K(1K)} & \multicolumn{3}{c}{COCO(5K)} & \multicolumn{3}{c}{Flickr30K(1K)} & \multicolumn{3}{c}{COCO(5K)} \\
& & $R@1$ & $R@5$ & $R@10$ & $R@1$ & $R@5$ & $R@10$ & $R@1$ & $R@5$ & $R@10$ & $R@1$ & $R@5$ & $R@10$ \\
\midrule
\multirow{7}{*}{COCO} & CLIP$_0$ & $\overline{16.9}$ & 37.0 & 46.2 & $\overline{14.7}$ & 34.2 & 47.0 & $\overline{12.0}$ & 30.0 & 41.0 & $\overline{10.6}$ & 29.6 & 41.0  \\
 & CLIP$_{ct}$ & 20.6 & 42.8 & 56.4 & 6.2 & 17.8 & 26.1 & 16.1 & 38.5 & 50.4 & 4.7 & 14.3 & 21.8 \\
\cline{2-14}
 & $\alpha = 10$ & 25.7 & 50.4 & 60.3 & 11.6 & 28.4 & 30.9 & 17.3 & 40.2 & 54.6 & 7.9 & 20.9 & 34.7  \\
 & $\alpha = 15 $ & 28.1 & 54.3 & 66.7 & 14.0 & 32.8 & 45.4 & 20.7 & 45.8 & 58.0 & 9.7 & 26.0 & 36.4 \\
 & $\alpha = 20 $ & \textbf{27.9} & \textbf{53.4} & \textbf{64.4} & \textbf{14.5} & \textbf{34.0} & \textbf{46.1} & \textbf{20.2} & \textbf{45.0} & \textbf{57.2} & \textbf{10.1} & \textbf{26.4} & \textbf{37.4} \\
 & $\alpha = 25$ & 26.6 & 52.8 & 62.3 & 14.5 & 34.8 & 46.7 & 20.2 & 44.7 & 57.0 & 10.0 & 27.7 & 38.1 \\
 & $\alpha = 30$  & 25.5 & 51.7 & 61.8 & 14.7 & 35.0 & 47.1 & 18.4 & 42.8 & 55.5 & 10.2 & 27.0 & 38.3 \\ %
\midrule
 COCO+F30K & CLIP$_{jt}$ & \underline{30.1} & 55.9 & 60.1 & \underline{16.1} & 38.1 & 51.9 & \underline{22.5} & 48.5 & 59.6 & \underline{11.7} & 30.9 & 42.7 \\
 
\bottomrule
\end{tabular}
\caption{The final multimodal retrieval performance of different $\alpha$ on continual CLIP$_{Mod-X}$ training in the Experiment \ref{experiment_A}.}
\label{alpha_modx}
\end{center}
\vspace{-0.5em}
\end{table*}

From the table, we can find that although different $\alpha$ affects the performance of the CLIP$_{Mod-X}$, \textbf{different $\alpha$ does not significantly affect the effectiveness of the Mod-X framework.} The performance of CLIP$_{Mod-X}$ is better than CLIP$_{ct}$ under different $\alpha$. As $\alpha$ increases, the CLIP$_{Mod-X}$ better maintains its retrieval ability on past COCO samples. The Image-Text R@1 and Text-Image R@1 on COCO(5K) remain around 14.5\% and 10.0\%. However, an excessively large $\alpha$ also limits the model's ability to fit new datasets. With the value of $\alpha$ increased from 20 to 30, the Image-Text R@1 and Text-Image R@1 of the CLIP$_{Mod-X}$ on the Flickr30k(1K) drops from 27.9\% and 20.2\% to 25.2\% and 18.4\%.

\subsection{The detailed performance of different training strategies at final training phase in Experiment \ref{experiment_A}}
\label{detailed_A}

\begin{table*}[!h]
\small
\begin{center}
\setlength\tabcolsep{1.8pt}
\begin{tabular}{c|l|ccc|ccc|ccc|ccc}
\toprule
\multirow{3}{*}{\shortstack[c]{Pretraining\\ Dataset}} & \multirow{3}{*}{\shortstack[c]{Model}} & \multicolumn{6}{c}{Image-Text Retrieval(\%)} & \multicolumn{6}{c}{Text-Image Retrieval(\%)} \\
& & \multicolumn{3}{c}{Flickr30K(1K)} & \multicolumn{3}{c}{COCO(5K)} & \multicolumn{3}{c}{Flickr30K(1K)} & \multicolumn{3}{c}{COCO(5K)} \\
& & $R@1$ & $R@5$ & $R@10$ & $R@1$ & $R@5$ & $R@10$ & $R@1$ & $R@5$ & $R@10$ & $R@1$ & $R@5$ & $R@10$ \\
\midrule
\multirow{4}{*}{COCO} & CLIP$_0$ & $\overline{16.9}$ & 37.0 & 46.2 & $\overline{14.7}$ & 34.2 & 47.0 & $\overline{12.0}$ & 30.0 & 41.0 & $\overline{10.6}$ & 29.6 & 41.0  \\
 & CLIP$_{ct}$ & 20.6 & 42.8 & 56.4 & 6.2 & 17.8 & 26.1 & 16.1 & 38.5 & 50.4 & 4.7 & 14.3 & 21.8 \\
 & CLIP$_{EWC}$ & 22.2 & 43.1 & 57.0 & 6.1 & 17.2 & 26.5 & 17.0 & 39.1 & 51.2 & 4.5 & 13.9 & 22.0 \\
 & CLIP$_{Mod-X}$ & \textbf{27.9} & \textbf{53.4} & \textbf{64.4} & \textbf{14.5} & \textbf{34.0} & \textbf{46.1} & \textbf{20.2} & \textbf{45.0} & \textbf{57.2} & \textbf{10.1} & \textbf{26.4} & \textbf{37.4} \\
\midrule
COCO+F30K & CLIP$_{jt}$ & \underline{30.1} & 55.9 & 60.1 & \underline{16.1} & 38.1 & 51.9 & \underline{22.5} & 48.5 & 59.6 & \underline{11.7} & 30.9 & 42.7 \\ %
\bottomrule
\end{tabular}
\caption{The final multimodal retrieval performance of the different continual CLIP training strategies in the Experiment \ref{experiment_A}.}
\label{table:modx_f30k}
\end{center}
\vspace{-0.5em}
\end{table*}

From the results in the Table \ref{table:modx_f30k}, it is clear that our method CLIP$_{Mod-X}$
maintains its multimodal retrieval results on COCO(5K) after completing continual training on Flickr30K. The gap between CLIP$_0$ and CLIP$_{Mod-X}$ is just 0.2\% points in image-text retrieval $R@1$ and 0.5\% points in text-image retrieval $R@1$ on COCO(5K). At the same time, the retrieval results of the CLIP$_{Mod-X}$ on the test set Flickr30K(1K) are also affected by the training domain and have a significant increase. The $R@1$ performance of the CLIP$_{Mod-X}$ in image-text retrieval rise from 16.9\% (in CLIP$_0$) to 27.9\%. And the $R@1$ results in text-image retrieval increase from 12.0\% (in CLIP$_0$) to 20.2\%. The performance gap between CLIP$_{Mod-X}$ and CLIP$_{jt}$ on the Flickr30K is only at most 2.3\% points. Conversely, due to the model's spatial disorder in continual training, the performance of CLIP$_{ct}$ on COCO(5K) drops significantly. In addition, although the performance of CLIP$_{ct}$ on Flickr30K(1K) has improved, it is still far from the upper bound CLIP$_{jt}$. From the above experimental results, although CLIP$_{EWC}$ improves the accuracy of continual CLIP training on Flickr30K(1K), it does not preserve the model's understanding in past samples (COCO(5K)). According to the above comparisons, we can conclude that our Mod-X framework can not only maintain the representation alignment on old samples during continual CLIP learning but also improve the model's fitting ability to the current training data domain.

\subsection{The detailed performance of different training strategies at final training phase in Experiment \ref{Experiments_ECommerce-T2I}}

In table \ref{modx_vit_ec}, we show the performance of different training strategies at final training phase in Experiment \ref{Experiments_ECommerce-T2I}. Comparing the CLIP$_{Mod-X}$'s $R@1$ and $R@5$ results with others in different datasets, we can find that CLIP$_{vit32}$ model that have not been trained on ECommerce-T2I dataset have poor multimodal retrieval capabilities on EC(5K) dataset (11.3\% and 10.1\%). When fine-tuning the CLIP$_{vit32}$ on ECommerce-T2I, the $R@1$ and $R@5$ performance of all training strategies improves. Different from other strategies, our Mod-X framework improves the model's multimodal retrieval ability to the current training data domain while maintaining its performance to the previous data domain (Flickr30K and COCO).

\begin{table*}[h]
\small
\begin{center}
\setlength\tabcolsep{1.8pt}
\begin{tabular}{l|cc|cc|cc|cc|cc|cc}
\toprule
 \multirow{3}{*}{\shortstack[c]{Model}} & \multicolumn{6}{c}{Image-Text Retrieval(\%)} & \multicolumn{6}{c}{Text-Image Retrieval(\%)} \\
 & \multicolumn{2}{c}{Flickr30k(1K)} & \multicolumn{2}{c}{COCO(5K)} & \multicolumn{2}{c}{EC(5K)} & \multicolumn{2}{c}{Flickr30k(1K)} & \multicolumn{2}{c}{COCO(5K)} & \multicolumn{2}{c}{EC(5K)}\\
 & $R@1$ & $R@5$ & $R@1$ & $R@5$ & $R@1$ & $R@5$ & $R@1$ & $R@5$ & $R@1$ & $R@5$ & $R@1$ & $R@5$ \\
\midrule
  CLIP$_{vit32}$ & 77.7 & 94.5 & 50.1 & 74.6 & 11.3 & 27.6 & 58.9 & 83.5 & 30.2 & 55.6 & 10.1 & 25.5  \\
  CLIP$_{ct}$ & 63.4 & 87.2 & 36.8 & 61.5 & 16.6 & 40.7 & 44.4 & 71.0 & 20.6 & 42.6 & 15.8 & 40.5 \\
  CLIP$_{EWC}$ & 64.0 & 87.8 & 37.7 & 64.3 & 16.2 & 40.0 & 44.8 & 72.4 & 20.7 & 44.1 & 16.5 & 42.0 \\
  CLIP$_{Mod-X}$ & \textbf{73.1} & \textbf{92.1} & \textbf{47.1} & \textbf{70.5} & \textbf{20.1} & \textbf{44.8} & \textbf{55.6} & \textbf{79.9} & \textbf{27.9} & \textbf{51.0} & \textbf{20.0} & \textbf{44.8} \\
\midrule
  CLIP$_{ft}$ & 64.5 & 88.6 & 39.8 & 64.8 & 23.5 & 50.8 & 46.9 & 73.1 & 22.2 & 44.5 & 23.5 & 50.6 \\ %
\bottomrule
\end{tabular}
\caption{The final multimoal retrieval performance of the CLIP$_{ct}$, CLIP$_{Mod-X}$ and CLIP$_{ft}$ based on OpenAI's CLIP$_{vit32}$ on specific e-commerce dataset ECommerce-T2I (Experiment \ref{Experiments_ECommerce-T2I}).}
\label{modx_vit_ec}
\end{center}
\end{table*}

\subsection{The performance of the Mod-X when training in CC12M dataset}
\label{Experiment_C}
In this section, we show the performance of different continual training strategies in CC12M \cite{changpinyo2021conceptual} training dataset. The CC12M training dataset collects about 12M images and their raw descriptions harvested from the alt-text HTML attribute associated with the webscraped images, therefore representing a wider variety of content styles. Due to unavailable URLs, we utilize about 10M examples from this dataset. Firstly, we randomly and evenly split the CC12M dataset into 10 sub-datasets, each containing 1M image-text pairs. Then, we continuously train a CLIP based on these sub-datasets from scratch without any pre-training. \textbf{The purpose of this experiment is to demonstrate that our Mod-X framework still excels in large-scale continual pre-training.} In table \ref{cc12m_final}, we show the final retrieval performance of different continual training strategies in COCO(5K) and Flickr30K(1K) test sets. The CLIP$_{ct}$ means continual training without any other operations. The CLIP$_{Mod-X}$ means continual training using our Mod-X framework. And the CLIP$_{jt}$ refers to training CLIP model using the joint dataset CC12M.

\begin{table}[h]
\small
\begin{center}
\setlength\tabcolsep{1.8pt}
\begin{tabular}{l|ccc|ccc|ccc|ccc}
\toprule
 \multirow{3}{*}{\shortstack[c]{Model}} & \multicolumn{6}{c}{Image-Text Retrieval(\%)} & \multicolumn{6}{c}{Text-Image Retrieval(\%)} \\
 & \multicolumn{3}{c}{Flickr30k(1K)} & \multicolumn{3}{c}{COCO(5K)} & \multicolumn{3}{c}{Flickr30k(1K)} & \multicolumn{3}{c}{COCO(5K)} \\
 & $R@1$ & $R@5$ & $R@10$ & $R@1$ & $R@5$ & $R@10$ & $R@1$ & $R@5$ & $R@10$ & $R@1$ & $R@5$ & $R@10$ \\
\midrule
  CLIP$_{ct}$ & 35.50 & 64.80 & 76.10 & 17.38 & 39.24 & 51.68 & 24.54 & 49.96 & 61.44 & 12.10 & 29.60 & 40.26  \\
  CLIP$_{Mod-X}$ & \textbf{40.40} & \textbf{67.90} & \textbf{77.40} & \textbf{22.06} & \textbf{46.12} & \textbf{58.14} & \textbf{27.74} & \textbf{53.88} & \textbf{64.66} & \textbf{14.22} & \textbf{33.68} & \textbf{45.02} \\
\midrule
  CLIP$_{jt}$ & \underline{58.00} & 83.90 & 90.40 & \underline{34.38} & 60.30 & 71.50 & \underline{43.02} & 72.34 & 80.92 & \underline{22.63} & 46.44 & 58.35 \\ %
\bottomrule
\end{tabular}
\caption{The final multimoal retrieval performance of the CLIP$_{ct}$, CLIP$_{Mod-X}$ and CLIP$_{jt}$ on COCO(5K) and Flickr30K(1K).}
\label{cc12m_final}
\end{center}
\end{table}

\begin{figure}[!h]
    \centering
	\subfigure{
	\centering
	\includegraphics[width=3.2in]{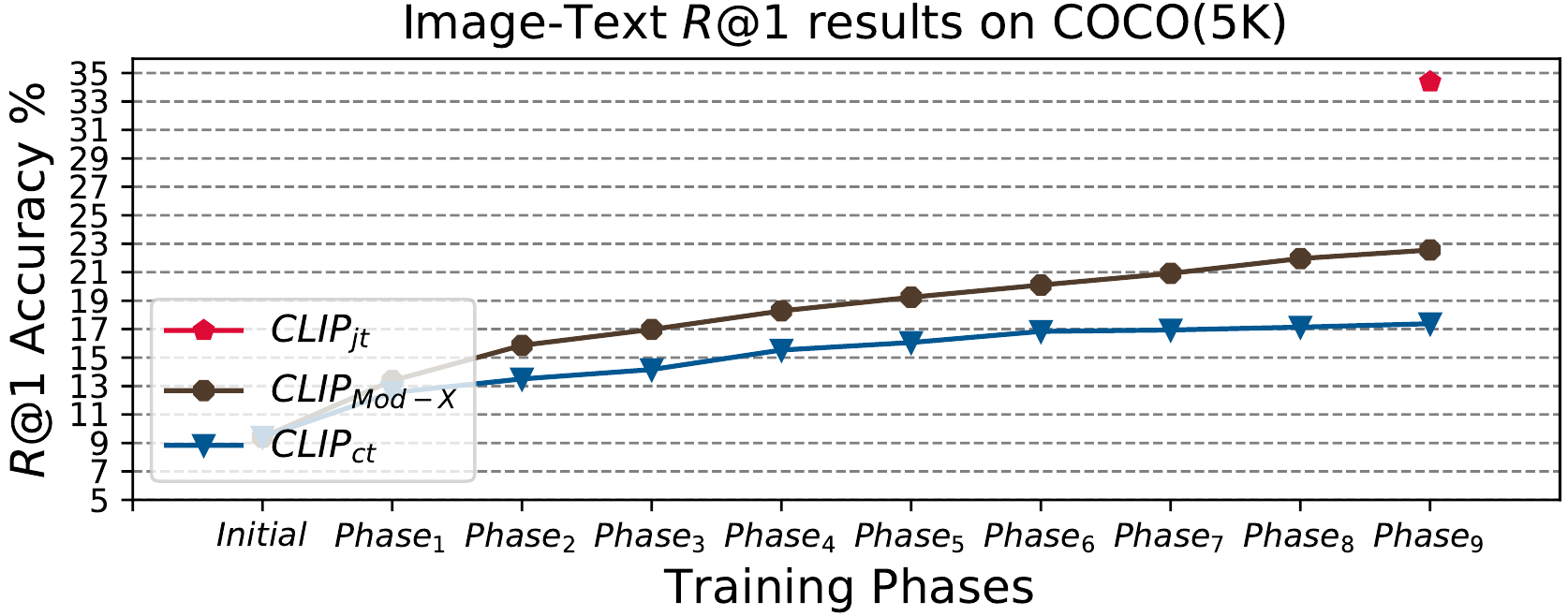}
	}
	\subfigure{
	\centering
	\includegraphics[width=3.2in]{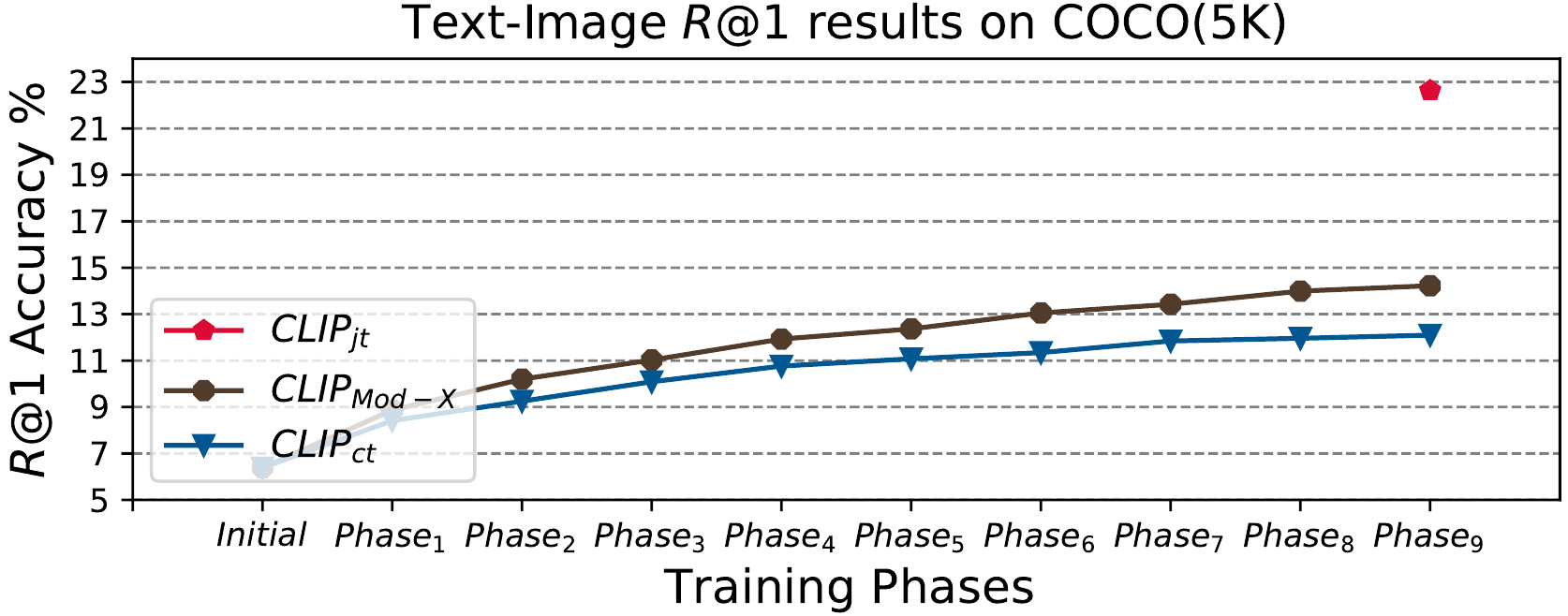}
	}
	
	\subfigure{
	\centering
	\includegraphics[width=3.2in]{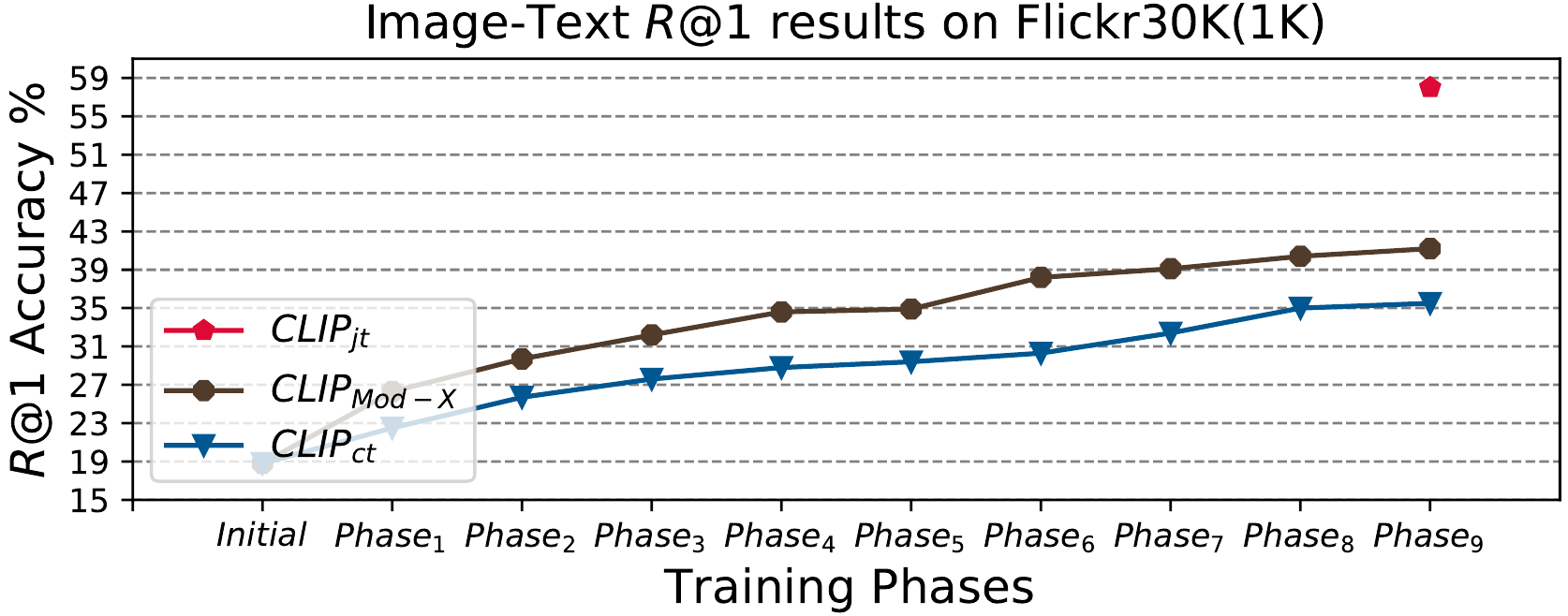}
	}
	\subfigure{
	\centering
	\includegraphics[width=3.2in]{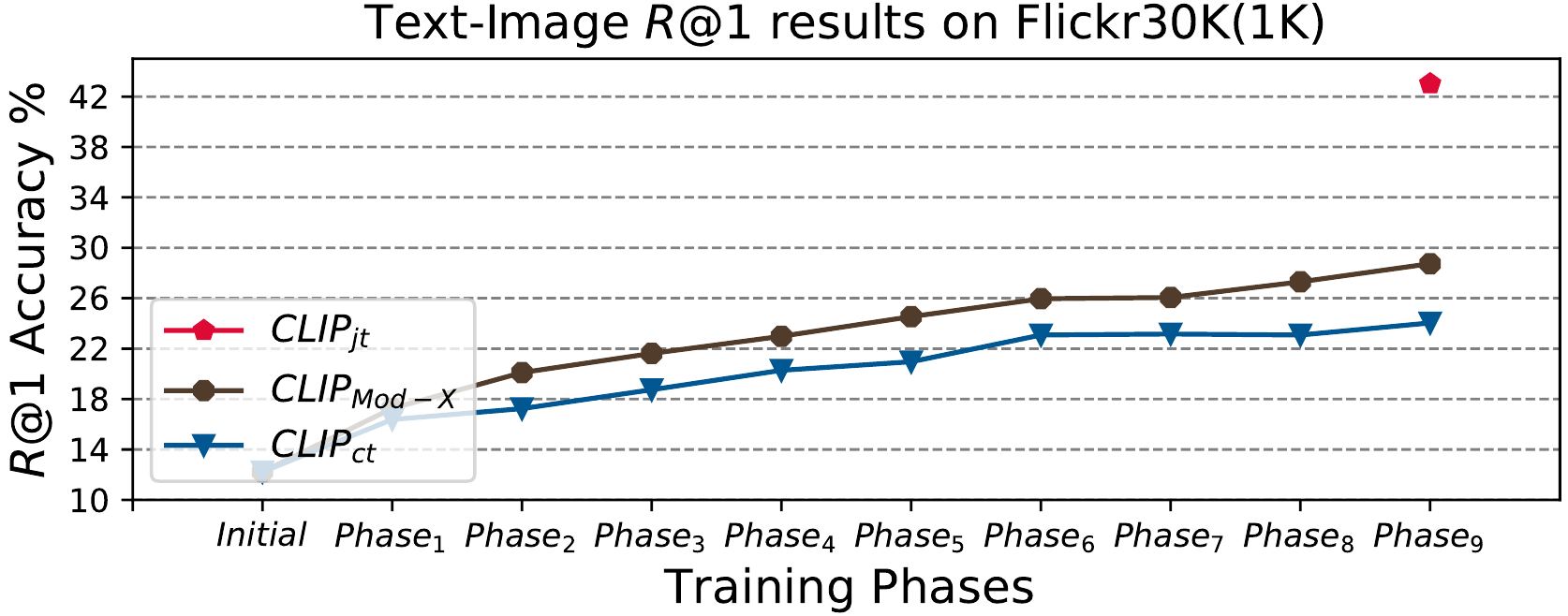}
	}
	\caption{The retrieval performance of different training strategies in each training phase on COCO(5K) and Flickr30K(1K).}
	\label{fig:modx_cc12m}
\end{figure}

Comparing the final performance of the three training strategies, Mod-X framework (CLIP$_{Mod-X}$) still outperforms CLIP$_{ct}$ in the large-scale pre-training. After continual pre-training, the CLIP$_{Mod-X}$ obtain 40.40\% Image-Text R@1 result and 27.74\% Text-Image R@1 result on Flickr30K(1K) test set, which surpasses the 35.50\% and 24.54\% of CLIP$_{ct}$. The results on COCO(5K) are similar to those on Flickr30K(1K). The Image-Text R@1 result of CLIP$_{Mod-X}$ on COCO(5K) is 4.68\% points higher than CLIP$_{ct}$ and the Text-Image R@1 result of CLIP$_{Mod-X}$ on COCO(5K) exceeds CLIP$_{ct}$ 2.12\% points. The detailed $R@1$ performance of three training strategies at each training phase can be seen in Figure \ref{fig:modx_cc12m}. 

Beside of this, we compare the performance of the Mod-X (CLIP$_{Mod-X}$), continual learning without other operations (CLIP$_{ct}$) and baseline joint learning (CLIP$_{jt}$) on linear probe top-1 accuracy (\%) and zero-shot image classification top-1 accuracy(\%) at final training phase. The results can be seen in the following Table \ref{table:modx_representation_quality}.

\begin{table*}[h]
\begin{center}
\setlength\tabcolsep{2pt}
\begin{tabular}{l|c|c|c|c|c|c|c}
\toprule
 \multirow{2}{*}{\shortstack[c]{Model}} & \multicolumn{6}{c}{Zero-Shot Image Classification(\%)} & Linear Probe(\%) \\
 \cline{2-8}
 & Cifar10 & Caltech101 & Places365 & ObjectNet & ImageNet & Average & ImageNet \\
\midrule
CLIP$_{jt}$ & 73.1 & 40.4 & 32.3 & 10.4 & 35.7 & 38.4 & 47.3 \\
\midrule
CLIP$_{Mod-X}$ & \textbf{71.2} & \textbf{35.8} & \textbf{28.7} & \textbf{8.3} & \textbf{29.8} & \textbf{34.8} & \textbf{41.6}\\
CLIP$_{ct}$ & 64.7 & 30.2 & 23.5 & 6.2 & 23.4 & 29.6 & 35.1 \\
\bottomrule
\end{tabular}
\caption{The linear probe top-1 accuracy (\%) and zero-shot image classification top-1 accuracy(\%) at final training phase of the CLIP$_{ct}$, CLIP$_{Mod-X}$ and CLIP$_{jt}$.}
\label{table:modx_representation_quality}
\end{center}
\end{table*}
From the results, we can find that the linear probe performance of CLIP$_{Mod-X}$ on ImageNet is significantly higher than that of CLIP$_{ct}$. This shows that the representation quality of the model continuously trained by the Mod-X framework (CLIP$_{Mod-X}$) is better than that of pure continual training (CLIP$_{ct}$). Comparing the zero-shot top-1 average results of the model on multiple classification datasets, it can be found that the representation generalization performance of the CLIP$_{Mod-X}$ is also significantly better than that of CLIP$_{ct}$. All of this shows that our Mod-X framework indeed improves the representation space quality of the CLIP model during continual training, which provides a good baseline for future continual self-supervised pre-training works.

\newpage
\subsection{The performance of Mod-X when continual training the OpenAI's CLIP on COCO and Flickr30K dataset}
\label{vit32_coco_f30k}
We set the CLIP$_{vit32}$ as the initial model, which is consistant with experiment \ref{Experiments_ECommerce-T2I}, and divide the joint-dataset (COCO and Flickr30K) into five sub-datasets uniformly and randomly to simulate streaming data. Because the pre-training datasets of CLIP$_{vit32}$ are not available, we train CLIP$_{vit32}$ on the joint-dataset to get the model CLIP$_{ft}$ as an upper bound for the performance of continual training. We apply our framework Mod-X in this setting and compare the final multimodal retrieval results with CLIP$_{ct}$, which is just continual training without any other operations, in Table \ref{table:modx_vit}.

\begin{table*}[h]
\small
\begin{center}
\setlength\tabcolsep{1.8pt}
\begin{tabular}{l|ccc|ccc|ccc|ccc}
\toprule
 \multirow{3}{*}{\shortstack[c]{Model}} & \multicolumn{6}{c}{Image-Text Retrieval(\%)} & \multicolumn{6}{c}{Text-Image Retrieval(\%)} \\
 & \multicolumn{3}{c}{Flickr30k(1K)} & \multicolumn{3}{c}{COCO(5K)} & \multicolumn{3}{c}{Flickr30k(1K)} & \multicolumn{3}{c}{COCO(5K)} \\
 & $R@1$ & $R@5$ & $R@10$ & $R@1$ & $R@5$ & $R@10$ & $R@1$ & $R@5$ & $R@10$ & $R@1$ & $R@5$ & $R@10$ \\
\midrule
  CLIP$_{vit32}$ & $\overline{77.7}$ & 94.5 & 98.3 & $\overline{50.1}$ & 74.6 & 83.0 & $\overline{58.9}$ & 83.5 & 90.1 & $\overline{30.2}$ & 55.6 & 66.7  \\
  CLIP$_{ct}$ & 85.6 & 97.3 & 98.8 & 59.7 & 83.2 & 90.2 & 71.2 & 91.5 & 94.9 & 43.5 & 70.9 & 80.6 \\
  CLIP$_{Mod-X}$ & \textbf{86.9} & \textbf{97.7} & \textbf{99.3} & \textbf{62.1} & \textbf{85.6} & \textbf{91.7} & \textbf{73.4} & \textbf{92.9} & \textbf{96.2} & \textbf{46.2} & \textbf{73.5} & \textbf{82.6} \\
\midrule
  CLIP$_{ft}$ & \underline{86.3} & 97.2 & 99.1 & \underline{63.6} & 86.4 & 92.3 & \underline{72.7} & 92.6 & 96.3 & \underline{46.3} & 73.1 & 82.3 \\ %
\bottomrule
\end{tabular}
\caption{The final multimoal retrieval performance of the CLIP$_{ct}$, CLIP$_{Mod-X}$ and CLIP$_{ft}$ based on OpenAI's CLIP$_{vit32}$ with VIT-B/32 vision encoder.}
\label{table:modx_vit}
\end{center}
\end{table*}

The performance of our framework Mod-X is still better than CLIP$_{ct}$ on all of the evaluation settings. Comparing the $R@1$ results on the test set Flickr30K(1K), we can find that CLIP$_{Mod-X}$ not only surpasses the initial results (CLIP$_{vit32}$) but also 1.3\% points and 2.2\% points higher than CLIP$_{ct}$. The results on COCO(5K) also illustrate that our framework not only resists the cognitive disorder of the model but also fits the new data domain better than CLIP$_{ct}$. The $R@1$ results of CLIP$_{Mod-X}$ on COCO(5K) surpasses the CLIP$_{ct}$ by 2.4\% and 2.7\% points, respectively.

\end{document}

%% file: content/1_abs.tex
\begin{abstract}
Large-scale multi-modal contrastive learning frameworks like CLIP typically require a large amount of image-text samples for training. However, these samples are always collected continuously in real scenarios. This paper discusses the feasibility of continual CLIP training using streaming data. Unlike continual learning based on self-supervised learning methods for pure images, which is empirically robust against catastrophic forgetting, CLIP’s performance degeneration in the continual setting is significant and non-neglectable.
By analyzing the changes in the model's representation space during continual CLIP training from a spatial geometry perspective, we explore and summarize these spatial variations as \textbf{Spatial Disorder (SD)}, which can be divided into \textbf{Intra-modal Rotation} and \textbf{Inter-modal Deviation}. 
Moreover, we empirically and theoretically demonstrate how SD leads to a performance decline for CLIP on cross-modal retrieval tasks. To alleviate SD, we propose a new continual vision-language representation learning framework \textbf{Mod-X}: \textbf{M}aintain \textbf{o}ff-\textbf{d}iagonal information-matri\textbf{X}. By selectively aligning the off-diagonal information distribution of contrastive matrices, the Mod-X improves the capability of the multi-modal model by maintaining the multi-modal representation space alignment on the old data domain during continuously fitting the new training data domain. Experiments on commonly used datasets with different scales and scopes have demonstrated the effectiveness of our method.

\end{abstract}

%% file: content/2_intro.tex
\section{Introduction}
Recently, multi-modal pre-trained models such as CLIP \cite{radford2021learning} have attracted much attention. By utilizing these pre-trained models, many works have achieved new progress in downstream tasks such as image classification, semantic segmentation, object detection, speech recognition \cite{wei2022mvp,wang2021dense,xie2021detco,baevski2020wav2vec}, etc. Although the CLIP model has strong generalization in open-world data, as mentioned in its original paper \cite{radford2021learning}, the ability to match image-text samples that are not in its training data distribution is still weak. The natural idea to alleviate this problem is to scale up the training data that covers different data domains. However, it is impractical to train infinite data distribution with limited hardware resources at once. 

To address the above problems, this paper mainly explores the feasibility of continuously training the CLIP model through streaming data, a training paradigm that follows Continual Learning (CL) \cite{mccloskey1989catastrophic}. Traditional supervised continual learning has been proven to suffer from catastrophic forgetting \cite{rebuffi2017icarl,kirkpatrick2017overcoming}: The model's performance on old tasks drops significantly as training phases rising. Recently, some works \cite{ni2021self,hu2021well} have validated that self-supervised models based on pure images like SimCLR \cite{chen2020simple} and BarlowTwins \cite{zbontar2021barlow} do not suffer from severe catastrophic forgetting during continual training. Some works \cite{madaan2021representational,thai2021does} conjecture that the reason is that the contrastive loss is not directly affected by the supervised signal, and the self-supervised framework does not have a Softmax function to amplify the influence of labels.

However, the performance of CLIP with a continual training setting is clearly different from the self-supervised continual training, which only uses images, though they both utilize contrastive loss. There is a significant degradation of multi-modal retrieval results with continual CLIP training compared with joint training (the experiment results are shown in Section \ref{explore} and \ref{experiments}). 
By analyzing the changes in the model's representation space during continual CLIP training from a spatial geometry perspective,
we explore and summarize these spatial variations as \textbf{Spatial Disorder (SD)}, which can be divided into \textbf{Intra-modal Rotation} and \textbf{Inter-modal Deviation}. The intra-modal rotation represents the representation space of the single-modal feature extractor (vision or language) within the CLIP rotates around the center of the high-dimensional sphere.
The inter-modal deviation represents the shift of representation alignment of different modal extractors (vision and language) to the same entities during continual training. Moreover, we demonstrate how intra-modal rotation and inter-modal deviation lead to a performance decline for CLIP on cross-modal retrieval tasks in both empirically and theoretically.

To alleviate this SD in continual CLIP training, we 
propose a simple yet effective framework Mod-X: \textbf{M}aintain \textbf{o}ff-\textbf{d}iagonal information-matri\textbf{X}. Unlike contrastive loss \cite{oord2018representation} only focuses on widening the similarity gap between positive and negative sample pairs, the Mod-X framework pays more attention to representation space alignment. The elements in the contrastive matrix represent the similarity between visual and textual entities, which also refer to the included angle between visual and textual representation vectors when the length of vectors is \textbf{1}. The angle distribution between the vectors represents the inherent representation space structure of the model under the current samples. 
By selectively aligning the distribution of the off-diagonal elements, Mod-X preserves the spatial relationships between modals of various old entities while fitting the current vision-language data during continual training.
The experiments (in Section \ref{method}, \ref{experiments} and Appendix \ref{Appendix_to_experiments}) on commonly used datasets with different scales and scopes show that our Mod-X framework improves the capability of the multi-modal model by maintaining the multi-modal representation space alignment on the old data domain during continuously fitting the new training data domain. 
The contributions of this paper are summarized as follows:
\begin{itemize}
\item We discuss the feasibility of training the CLIP model continuously through streaming data. Empirical experiments demonstrate that continual CLIP training leads to persistent performance degrades on cross-modal retrieval tasks, which is  different from the phenomenon of continual learning based on self-supervised learning methods for pure images.

\item We explore and summarize the model's spatial variation during continual CLIP training as Spatial Disorder, which can be divided into intra-modal rotation and inter-modal deviation. Furthermore, we demonstrate how spatial disorder leads to a performance decline for CLIP on cross-modal retrieval tasks in both empirically
and theoretically (in Section \ref{explore}).

\item We propose a simple yet effective continual CLIP training framework \textbf{Mod-X} that alleviates space disorder during continual CLIP training by selectively aligning contrastive matrices' off-diagonal information. Experiments (in Section \ref{experiments} and Appendix \ref{Appendix_to_experiments}) on commonly used datasets with different scales and scopes have evaluated the effectiveness of our method.
\end{itemize}






%% file: content/3_relate.tex
\section{Related Work}
\textbf{Continual Learning.} Continual learning (CL) \cite{thrun1995lifelong}, or incremental learning, mainly focuses on supervised tasks. In addition to the vision-based tasks \cite{de2021continual,kj2021incremental,cha2021ssul,ahn2021ss}, some works discuss language-based tasks \cite{biesialska2020continual,sun2019lamol}. We can summarize the existing continual learning methods into three categories: regularization \cite{kirkpatrick2017overcoming,ahn2019uncertainty,ni21revisit}, replay \cite{rebuffi2017icarl,rolnick2019experience,wang2021ordisco}, and architecture \cite{thai2021does,ni2021self,hu2021well,madaan2021representational}. However, traditional supervised continual learning methods are limited by labels and unsuitable for self-supervised or unsupervised situations.

\quad In unsupervised and self-supervised single-modal continual training, the latest work \cite{thai2021does,ni2021self,hu2021well,madaan2021representational} has drawn some conclusions different from those of supervised continual learning. However, only a few pieces \cite{srinivasan2022climb,fan2022unified} focus on incremental multi-modal learning. However, \cite{srinivasan2022climb} did not propose a new method to alleviate the catastrophic forgetting problem in multimodal continual learning. Instead, it provided baselines for the state-of-the-art supervised unimodal continual learning methods when applied to some single-model multimodal tasks. And \cite{fan2022unified} discussed continuous updates in multimodal graphs, which is far from our intended goal of continuously updating multimodal pre-trained models. Because of the cooperation between different modalities, continual multi-modal pre-training shows different performance and complex problems from single-modal continual training.


\textbf{Visual-Language Representational Learning.} Vision-language representation learning based on contrastive loss \cite{oord2018representation}, such as CLIP \cite{radford2021learning}, has attracted a lot of attention in various fields \cite{radford2021learning,li2021supervision,andonian2022robust}. And the pre-trained model performs surprisingly well on downstream tasks \cite{shu2022test,wang2022learning,chowdhury2022novelty}. At the same time, the large-scale image-text datasets, e.g., Laion400M \cite{schuhmann2021laion} and Conceptual Captions \cite{sharma2018conceptual}, have played a key role in multimodal pre-training. Although large-scale open-world datasets contain various samples, the pre-trained model still loses the ability to perfectly match image-text sample pairs that are not in its training data domain \cite{radford2021learning}. 





%% file: content/4_explori.tex
\begin{figure*}[t]
    \centering
	\subfigure{
	\centering
	\includegraphics[width=3.2in]{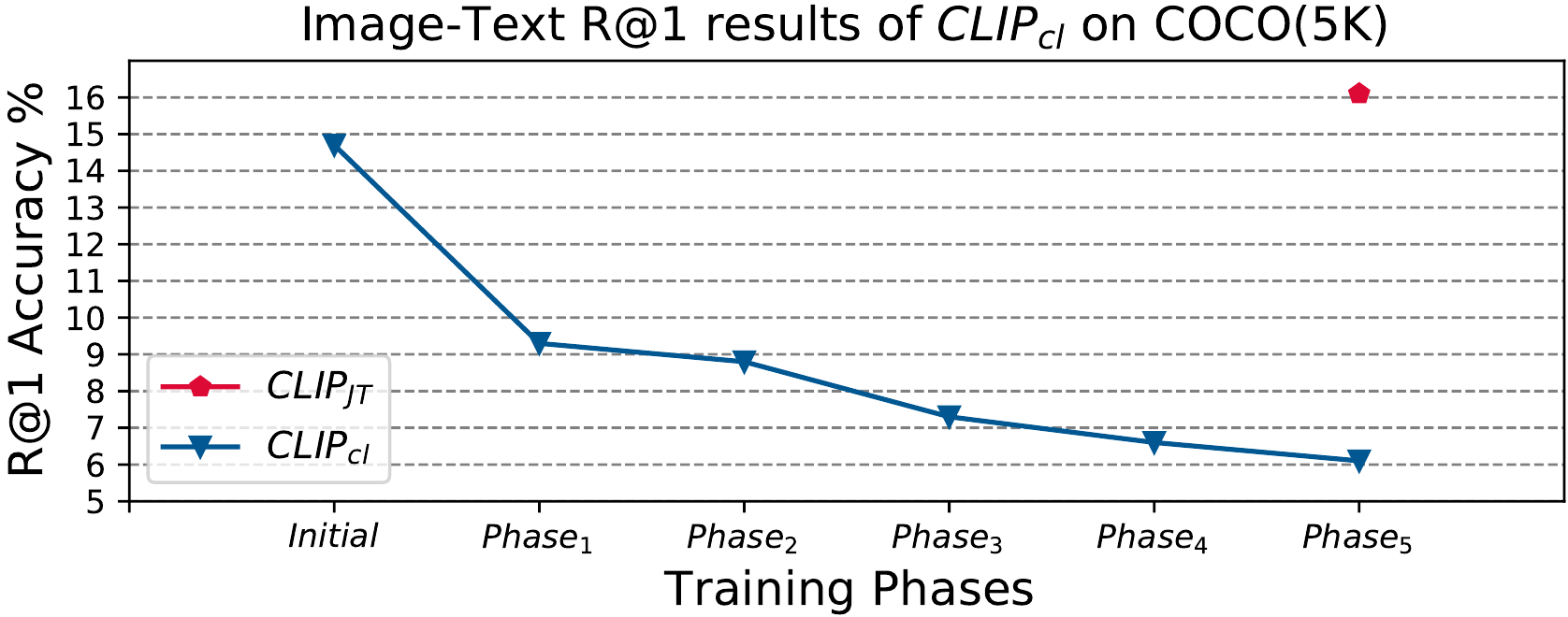}
	}
	\subfigure{
	\centering
	\includegraphics[width=3.2in]{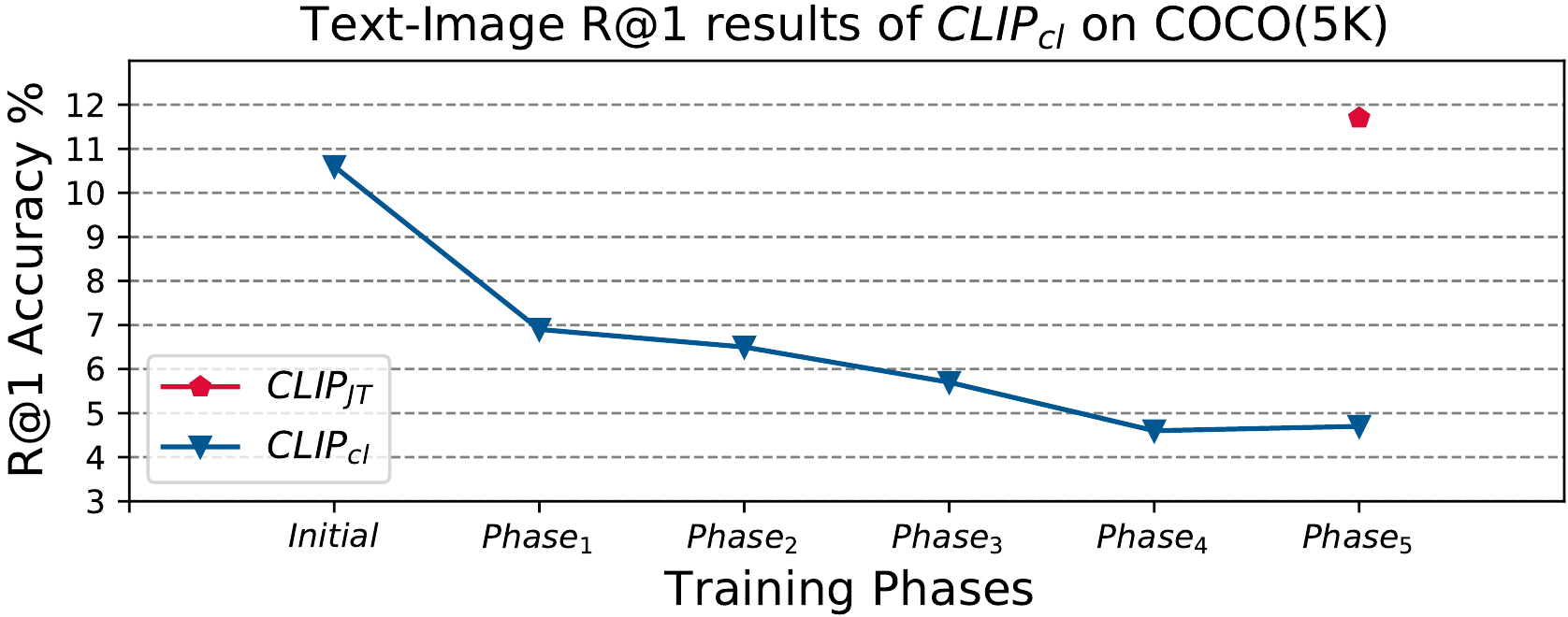}
	}
    \vspace{-.5em}
	
	\subfigure{
	\centering
	\includegraphics[width=3.2in]{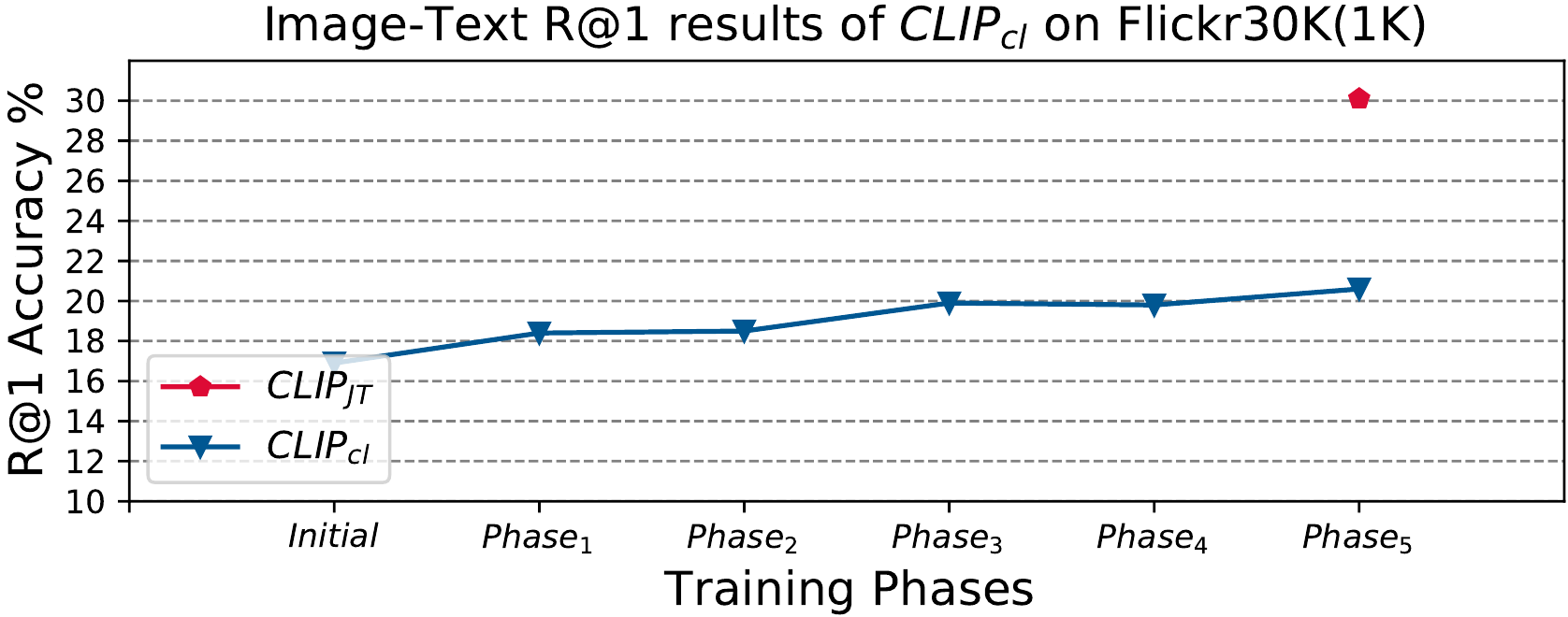}
	}
	\subfigure{
	\centering
	\includegraphics[width=3.2in]{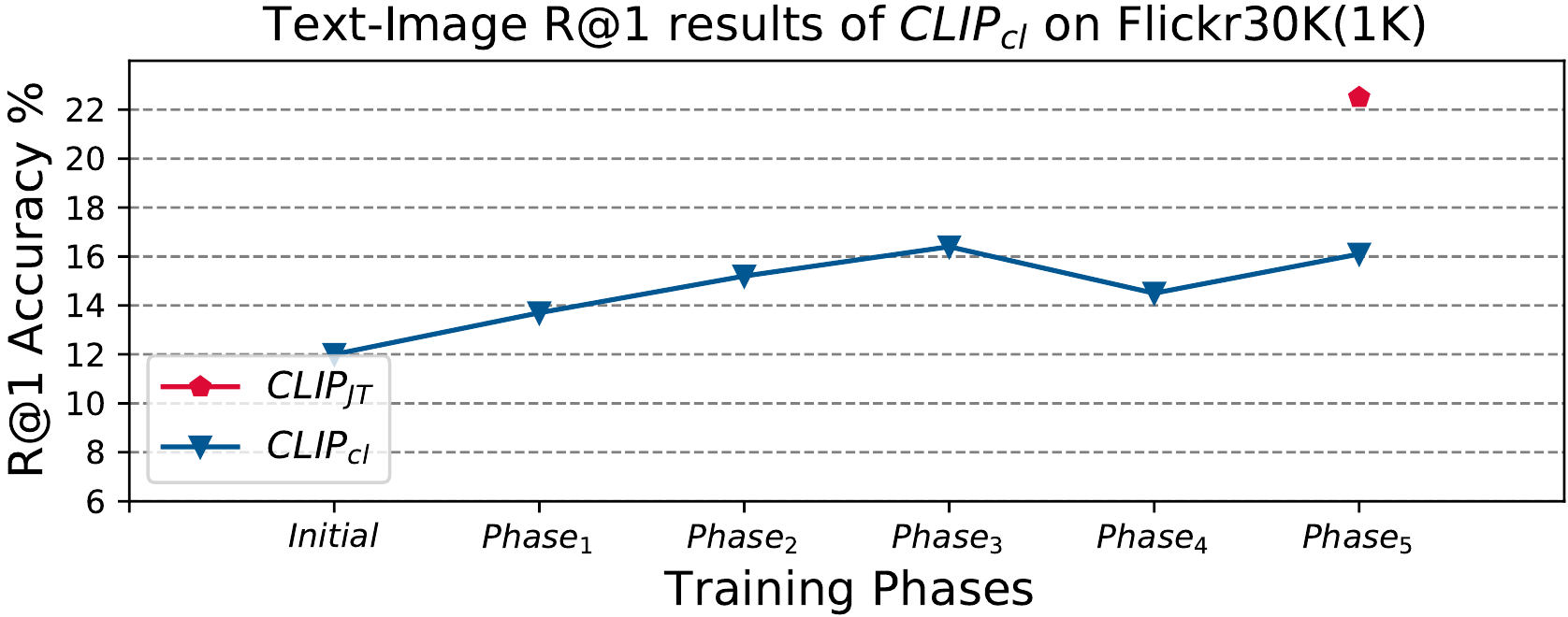}
	}
	\vspace{-1em}
	\caption{The multi-modal retrieval $R@1$ results of CLIP$_t$ $(0\leq t\leq 5)$ on test sets COCO (5K) and Flickr30k (1K). The two sub-figures on the left show the Image-Text retrieval $R@1$ performance of CLIP$_t$ on the continual training phase $t$. The initial training phase represents the performance of CLIP$_{0}$. The rights show the Text-Image $R@1$ results of CLIP$_t$ on the continual training phase $t$. The pentagon points (CLIP$_{jt}$) show the results of the CLIP under joint training, which is an upper bound for continual CLIP training (CLIP$_{ct}$).
	}
	\vspace{-.5em}
	\label{fig:base_retrival}
\end{figure*}
\section{Spatial Disorder in Continual CLIP}
\label{explore}
This section mainly aims to explore the characteristics of the CLIP model while training continually. By analyzing the changes in the model’s representation space from a spatial geometry perspective during continual CLIP training, we explore and summarize these spatial variations as Spatial Disorder (SD), which can be divided into intra-modal rotation and inter-modal deviation. Then, we demonstrate how intra-modal rotation and inter-modal deviation lead to a performance decline for CLIP on cross-modal retrieval tasks in both empirically and theoretically, respectively.

\textbf{Exploration Setup.} 
To ensure the controllability of the exploration, we train a CLIP$_0$ model from scratch on the COCO dataset \cite{lin2014microsoft} based on the OpenAI source code \cite{OpenAi} and use it as the initial state (start) of continual CLIP training. After that, we divide the Flickr30K dataset \cite{young2014image} into five sub-datasets \{$D_1$,$D_2$,$D_3$,$D_4$,$D_5$\} uniformly and randomly to simulate streaming data. Then we train the CLIP$_0$ based on these sub-datasets sequentially. We name this pure continual training without other operations as CLIP$_{ct}$.
After finishing five training phases, we obtain the model CLIP$_5$. For comparison with CLIP$_{5}$, we joint training a CLIP$_{jt}$ model using joint dataset COCO and Flickr30K, as the upper bound of the CLIP$_{5}$. The hyper-parameters for all training phases are kept the same, and detailed settings of CLIP model and training hyper-parameters can be seen in Appendix \ref{Experiment_Setting}. 



\subsection{The Performance of Continual CLIP Training}
\label{explore_results}

We show the $R@1$ retrieval results of CLIP$_t$ $(0\leq t\leq 5)$ on the test set COCO(5K) and Flickr30K(1K) in Figure \ref{fig:base_retrival}. By comparing the performances of the CLIP$_0$ (initial phase) and CLIP$_{jt}$ on Flickr30K(1K), we can find that the retrieval performance of CLIP$_{jt}$ (red point) is significantly better than that of CLIP$_0$ (initial) which is not trained on Flickr30K. This phenomenon shows that \textbf{the performance of the CLIP model is affected by the training data domain}, which is consistent with the conclusion of the paper \cite{radford2021learning}. Besides this, it can be clearly seen that the multi-modal retrieval performance of the CLIP$_{ct}$ on the COCO(5K) declines continually with the rising of training phases. The final Image-Text R@1 result of CLIP$_5$ on COCO(5K) plummeted from the initial 14.7\% to 6.1\%, and the Text-Image results dropped from 10.6\% to 4.7\%. The gap with CLIP$_{jt}$ reached 10.0\% and 7.0\%, respectively. 
On the other hand, CLIP$_{ct}$ exhibits a slow and erratic increase in multi-modal retrieval $R@1$ results on the test set Flickr30K(1K). Although the results between CLIP$_{ct}$ and CLIP$_{jt}$ on the Image-Text $R@1$ has been narrowed from the original 13.2\% to 9.5\% while the Text-Image $R@1$ of CLIP$_{ct}$ has increased from 12.0\% to 16.1\%, the gap between CLIP$_5$ and CLIP$_{jt}$ is still great. 

%% file: content/5_reason.tex
\subsection{The reasons for catastrophic forgetting}

In CLIP, the vision and language encoders normalize the final representation vector to a unit vector of length 1 using a dimension-based $L_2$ norm. 
This design makes the representation space in vision and language encoders form a high-dimensional unit sphere, respectively. Therefore, we ignore the influence of the representation vectors' length and track their direction changes. We summarize these spatial variations as Spatial Disorder (SD), which can be divided into intra-modal rotation and inter-modal deviation.  


\subsubsection{The Intra-modal Rotation}
\label{ImR}
\begin{figure}[h]
\subfigure[]{
\begin{minipage}[r]{0.5\linewidth}
\centering
\includegraphics[width = \linewidth]{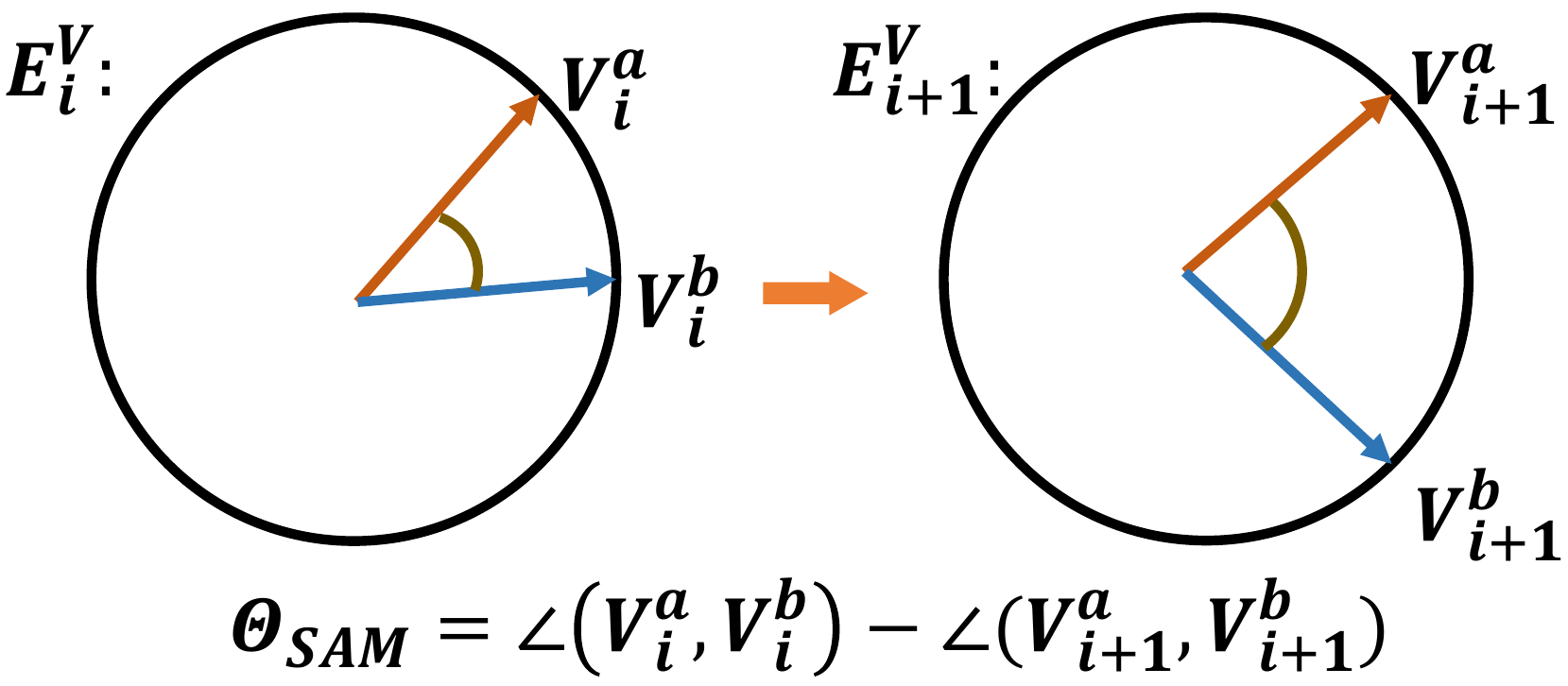}
\label{SAM_a}
\end{minipage}
}
\subfigure[]{
\begin{minipage}[l]{0.44\linewidth}
\centering
\includegraphics[width = \linewidth]{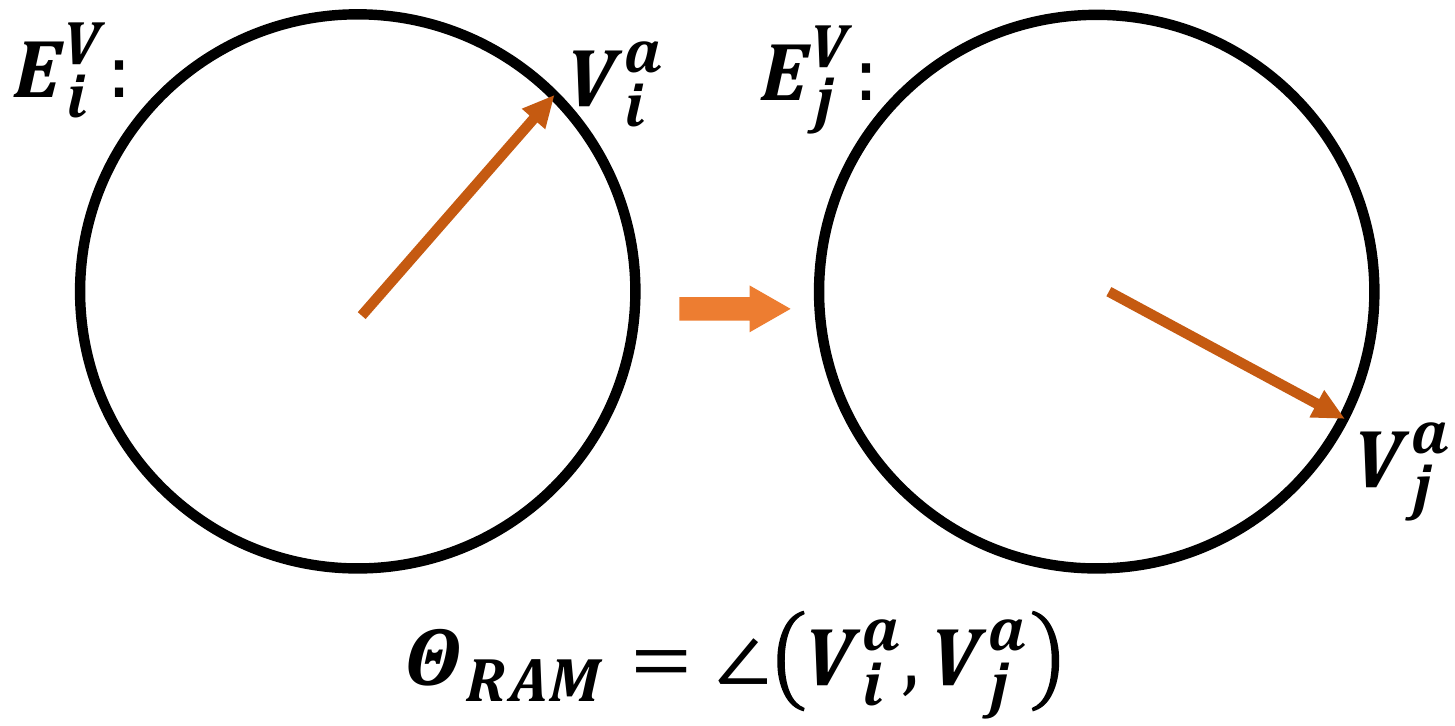}
\label{RAM_a}
\end{minipage}
}

\subfigure[]{
\setlength\tabcolsep{1pt}
\begin{minipage}[c]{1\linewidth}
\footnotesize
\begin{tabular}{|c|c|c|c|c|c|}
\toprule
 $\theta_{SAM} \in$ & $[0^{\circ}$, $5^{\circ}]$ & $(5^{\circ}$, $10^{\circ}]$ & $(10^{\circ}$, $15^{\circ}]$ & $(15^{\circ}$, $20^{\circ}] $ & $(20^{\circ}$, $180^{\circ}]$ \\
\midrule
$SAM_{0-1}$ &  54.23$\%$ & 31.64$\%$& 11.28$\%$ & 2.48$\%$ & 0.38$\%$ \\
$SAM_{1-2}$ &  61.18$\%$ & 30.21$\%$& 7.54$\%$ & 0.99$\%$ & 0.07$\%$ \\
$SAM_{2-3}$ &  61.44$\%$ & 30.00$\%$& 7.50$\%$ & 0.98$\%$ & 0.07$\%$ \\
$SAM_{3 - 4}$ &  55.33$\%$ & 31.75$\%$& 10.57$\%$ & 2.07$\%$ & 0.28$\%$ \\
$SAM_{4 - 5}$ &  50.51$\%$ & 32.14$\%$& 13.17$\%$ & 3.50$\%$ & 0.68$\%$ \\
$SAM_{0 - 5}$ &  42.94$\%$ & 31.12$\%$& 16.73$\%$ & 6.66$\%$ & 2.55$\%$ \\
\bottomrule
\end{tabular}
\label{SAM_b}
\end{minipage}
}
\caption{The sub-figure on the (a) shows a schematic diagram of computing $\theta_{SAM}$. The sub-figure on the (b) shows a schematic diagram of computing $\theta_{RAM}$. The table on the bottom (c) shows the distribution of the change of the included angle between any two samples in different training phases' vision representation space. And $SAM_{i-j}$ = $|SAM_i - SAM_j|$.}
\label{SAM}
\end{figure}

Firstly, we analyze the directional changes of the representation vectors of model's vision and language extractors during continual CLIP training. Taking the visual representation space as an example, we use the visual encoder $E^V_{i}$ in CLIP$_{i}$ to extract the image representations of the test set COCO(5K) and obtain the vision representation vectors sets $V_{i}=\{V_i^0,...,V_i^N,...,V_i^{5K}\}$, where $i=0,...,5 $ stands for five different training phases. After that, we take the inner product of each pair of vectors $<V_i^a,V_i^b>$, where $a$ and $b$ are arbitrary indexes in each vector set $V_{i}$ and perform $arccos$ operation, the inverse trigonometric function of cosine, to obtain their \textbf{S}elf-\textbf{A}ngle relationship \textbf{M}atrix ($SAM_{i}$). The $SAM_i^{(a,b)} = arccos(<V_i^a,V_i^b>)$.
Any element $SAM_i^{(a,b)}$ in the $SAM_i$ matrix represents the included angle between the sample $a$ and $b$ in the vision encoder $E^V_i$. By counting the difference value $\theta_{SAM} = \angle (V_i^a,V_i^b) - \angle (V_{i+1}^a,V_{i+1}^b)$ between the corresponding elements in two continual $SAM$ matrix $SAM_i$ and $SAM_{i+1}$ as shown in Figure \ref{SAM_a}, we get the following included angle change distribution in Figure \ref{SAM_b}.

From Figure \ref{SAM_b}, we can find that 80\% of the angle changes between any two vision representation vectors are between 0 and 10 degrees in continual training phases, while only 20\% are above 10 degrees. Moreover, less than 1\% of the angle changes are above 20 degrees. That angle changes between 15-20 degrees also only account for about 5\% of all image pairs. 
Therefore, we conclude that \textbf{the topology of the visual representation of the CLIP$_{ct}$ changes slowly during the continual CLIP training.} In Appendix \ref{representation_quality}, we reached the same empirical conclusion by comparing the representation quality of vision encoders.

In addition to discussing the change in the included angle between sample pairs in the visual representation space, by taking the inner product of the same sample's vision representation vector from different training phases' vision encoder $E_i^V$, we use the $arccos$ operation to compute the rotation angles $\theta_{RAM} = \angle (V_i^a,V_j^a)$ of each test sample $a$ in vision encoder $E_{i}^V$ and $E_{j}^V$ and get the \textbf{R}otation \textbf{A}ngle \textbf{M}atrix RAM$_{(i,j)}$. The $RAM_{(i,j)}^{a} = arccos(<V_i^a,V_j^a>)$. The schematic diagram can be seen in Figure \ref{RAM_a}. By counting the distribution of rotation angles, we get the following rotation angle distribution Table \ref{RAM}.

As shown in Table \ref{RAM}, we can find that the direction of the same sample in the visual representation space of different training phases has changed greatly. Only less than 0.4\% samples are rotated within 20 degrees in the continual CLIP training, while the samples rotated within 20-25 degrees are at most less than 9\%, and the samples of 25 degrees and above account for more than 90\%. We speculate that \textbf{the vision representation space of CLIP$_{ct}$ has undergone a large rotation around the high-dimensional sphere center during the continual training.} After analyzing the language representation space, we reach the same conclusion as the vision representation space. Detailed SAM and RAM of language encoders can be viewed in Appendix \ref{detail_language_encoder}.

According to our analysis of the geometric changes of the single-modal encoder's representation space during continual CLIP training, we conclude that: \textbf{During the continual CLIP training, the representation space in the CLIP$_{ct}$ is significantly rotated. The topology of the representation space is slightly rotated compared with the rotation of the whole representation space.} We name this phenomenon \textbf{Intra-modal Rotation}.

\begin{table}[h]

\begin{minipage}[c]{1\linewidth}
\footnotesize
\setlength\tabcolsep{0.5pt}
\begin{tabular}{|c|c|c|c|c|c|}
\toprule
 $\theta_{RAM} \in$ & $[0^{\circ}$, $15^{\circ}]$ & $(15^{\circ}$, $20^{\circ}]$ & $(20^{\circ}$, $25^{\circ}]$ & $(25^{\circ}$, $30^{\circ}] $ & $(30^{\circ}$, $180^{\circ}]$ \\
\midrule
$RAM_{(0,1)}$ &  0.00$\%$ & 0.20$\%$& 4.62$\%$ & 22.68$\%$ & 72.50$\%$ \\
$RAM_{(1,2)}$ &  0.00$\%$ & 0.40$\%$& 8.30$\%$ & 34.11$\%$ & 57.20$\%$ \\
$RAM_{(2,3)}$ &  0.00$\%$ & 0.30$\%$& 8.40$\%$ & 34.29$\%$ & 57.01$\%$ \\
$RAM_{(3,4)}$ &  0.00$\%$ & 0.00$\%$& 1.89$\%$ & 17.11$\%$ & 81.00$\%$ \\
$RAM_{(4,5)}$ &  0.00$\%$ & 0.00$\%$& 0.00$\%$ & 3.20$\%$ & 96.81$\%$ \\
$RAM_{(0,5)}$ &  0.00$\%$ & 0.00$\%$& 0.00$\%$ & 0.21$\%$ & 99.80$\%$ \\
\bottomrule
\end{tabular}
\end{minipage}
\caption{
The table on the bottom shows the rotation angle distribution of the same samples in different  training phases.}
\label{RAM}
\vspace{-1em}
\end{table}

\begin{figure*}[h]
	\centering
	\includegraphics[height=2.2in]{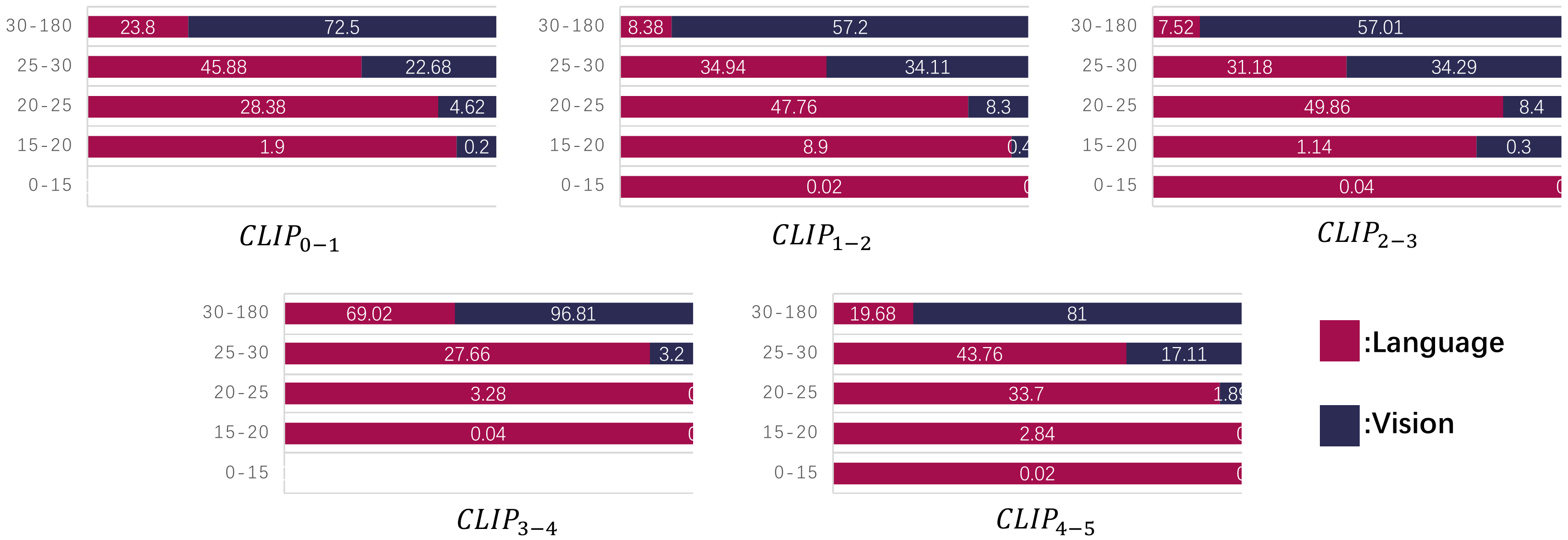}
	\caption{The comparison of the rotation distributions of the vision encoder and langugae encoder during continual CLIP training. CLIP$_{i-j}$ refers to the CLIP's continual training from training phase $i$ to $j$. The values under the same color represent the proportion of test samples to total samples in each rotation angle interval of the same modality.}
	\label{RAM_difference}
\end{figure*}
\subsubsection{The Inter-modal Deviation}
\label{ImD}
Although the topology of the single-modal representation space changes during continual training, this slight rotation should not be the main reason for the significant degradation of CLIP's multi-modal retrieval performance in continual training. To this end, we conduct a thought experiment: it is known that the representation spaces of vision and language encoders exhibit significant spatial rotations during continual training. Now we assume that the topology of the single-modal representation space is completely fixed during continual training. Therefore, if the CLIP$_{ct}$'s performance on multi-modal retrieval tasks does not degrade during continual training, \textbf{the rotations of the two encoders' representation spaces should be synchronized}. However, the fact is the \textbf{opposite}. So we think \textbf{there is a deviation between the rotation of the vision and language representation spaces}. Based on this suppose, we compare the rotation distributions of vision encoder (Table \ref{RAM}) and language encoder (Appendix \ref{detail_language_encoder}) and draw the rotation distribution comparison diagram (Figure \ref{RAM_difference}).
The values under the same color represent the proportion of test samples to total samples in each rotation angle interval of the same modality. Comparing the difference in the distribution of rotation angles of the vision and language encoders, we can see that the space rotations of the two encoders are very different in the continual training. The rotation of language representation space is mostly concentrated between 20-30 degrees, while the vision's rotations are mostly between 30-180 degrees. This shows that the rotation of the representation space of the two modal extractors within CLIP$_{ct}$ is not synchronized during the continual training, which verifies our previous deduction: \textbf{The unsynchronized rotation of the vision and language representation spaces leads to representation space deviations between the CLIP's modal encoders (vision and language).} We name this phenomenon \textbf{Inter-modal Deviation}.

\begin{figure}[!h]
\subfigure[]{
\begin{minipage}[c]{1\linewidth}
\centering
\includegraphics[width = \linewidth]{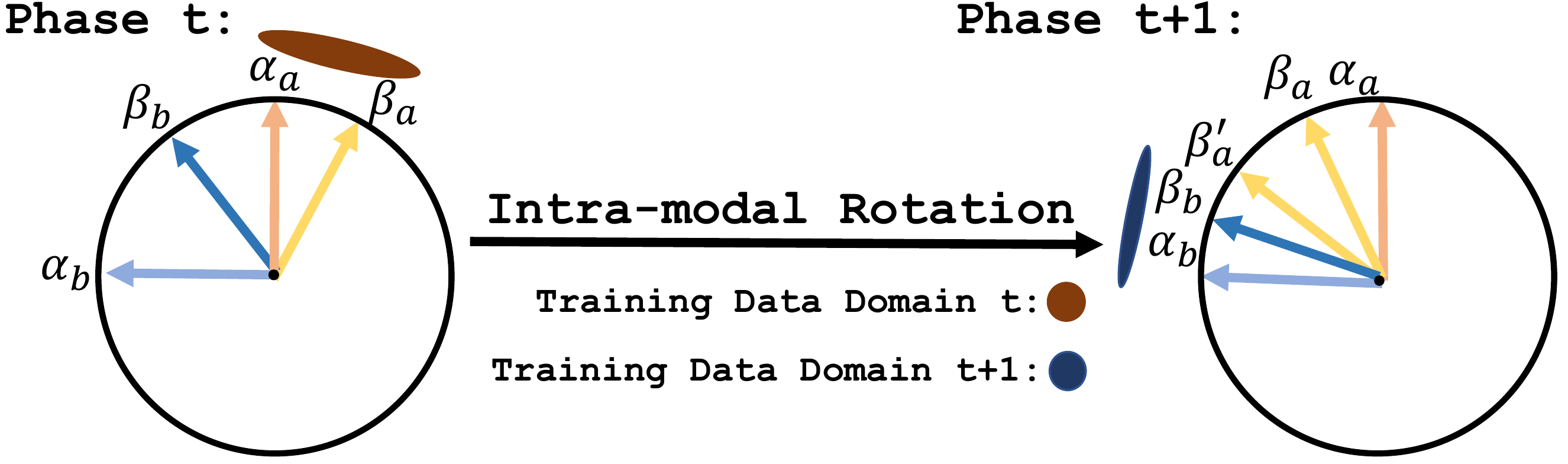}
\end{minipage}
\label{relation_rot}
}

\subfigure[]{
\begin{minipage}[c]{1\linewidth}
\centering
\includegraphics[width = \linewidth]{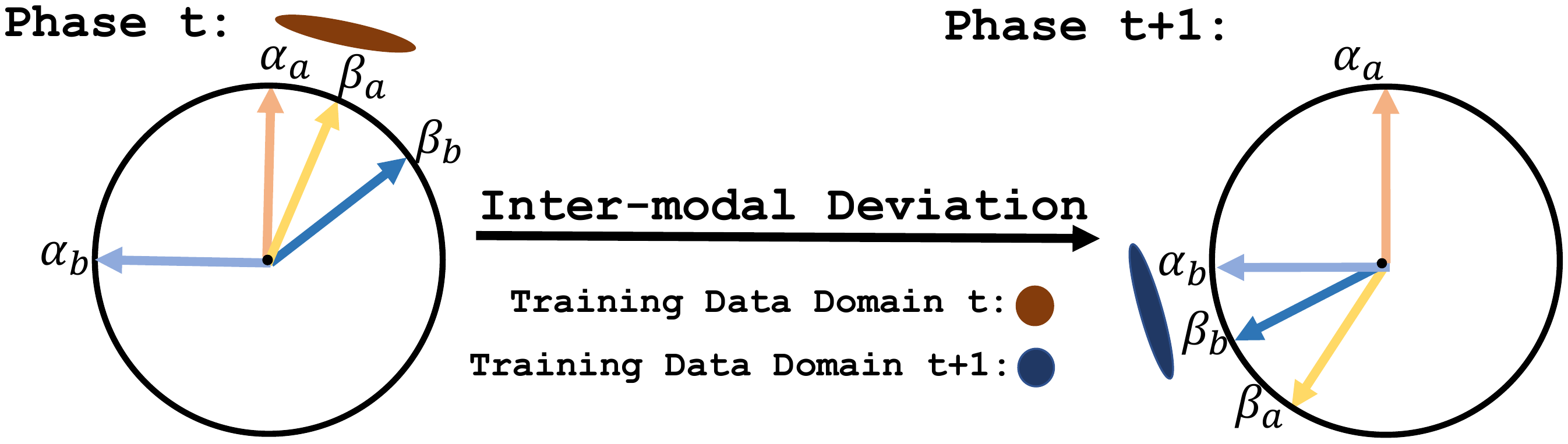}
\end{minipage}
\label{relation_dev}
}
\caption{The Schematic illustration of spatial disorder caused by intra-modal rotation and inter-modal deviation. The $\alpha$ is vision representation and $\beta$ is language representation. The $a$,$b$ denote different image-text samples.
}
\label{concept_indistinct}
\end{figure}
\begin{figure*}[!t]
	\centering
	\includegraphics[height=3.0in]{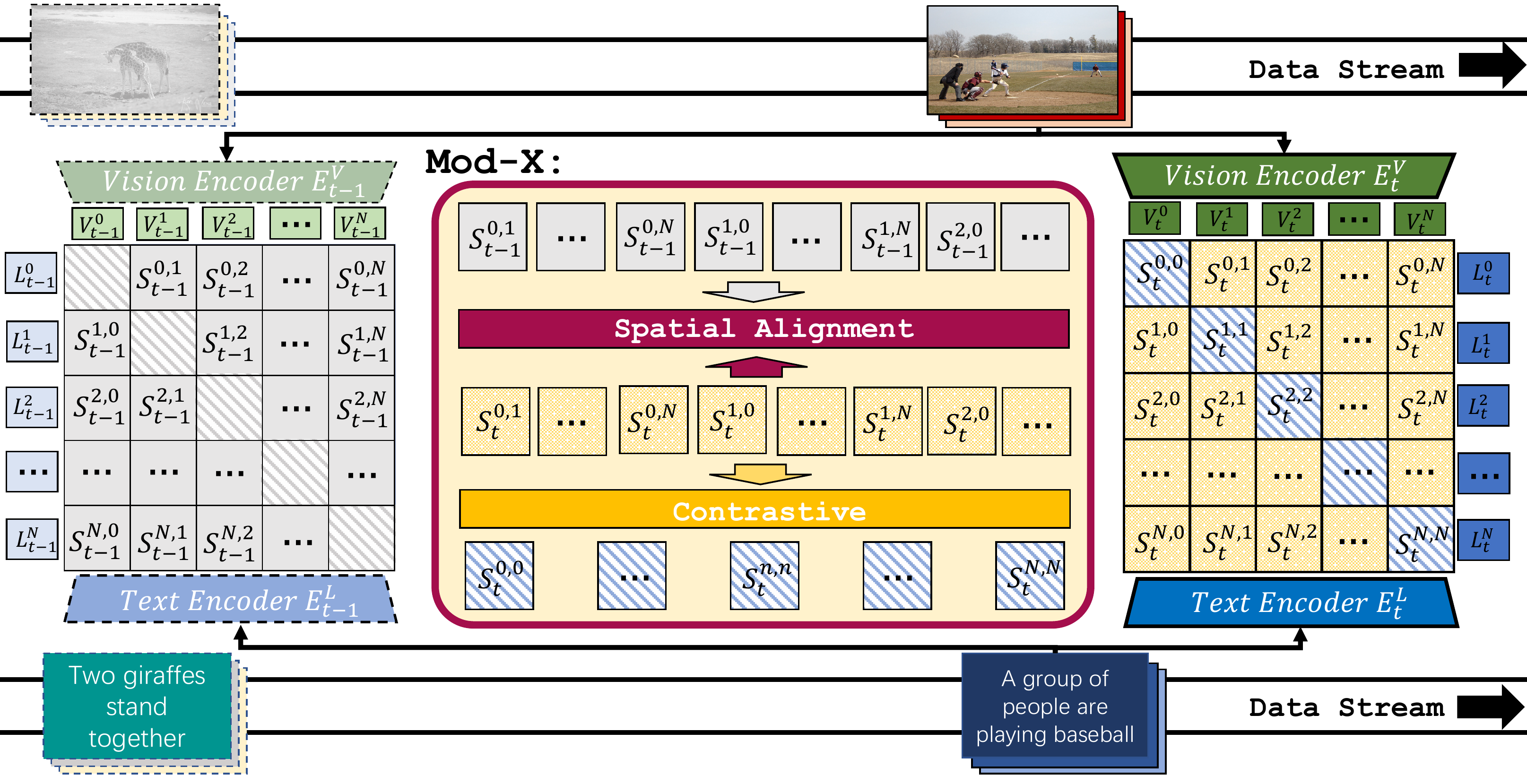}
	\caption{The Mod-X framework mainly consists of two sub-modules. Spatial Alignment helps the current model align the representation space of the old model based on current data. And Contrative helps the model fit the current training data domain. 
    }
	\label{main_process}
\end{figure*}
\subsubsection{The Relationship between Spatial Disorder and Contrastive Matrix}
\label{relationshape_sd_cm}
How do spatial disorder cause the model to mis-align the old sample's vision and language representation? We show a schematic here to illustrate this. As shown in Figure \ref{concept_indistinct}, the $\alpha$ is vision representation and $\beta$ is language representation. The $a$,$b$ denote different image-text samples. 
For the convenience of illustration, we fix the vision vectors' relative location and rotate the language vectors to represent the unsynchronous rotation of the two modal spaces. 
When intra-modal rotation happens (Figure \ref{relation_rot}), $\beta_a$ in training phase $t+1$ is rotated to $\beta'_a$, the modal similarity between $a$ and $b$ shift from ($\beta_a^T \alpha_a > \beta_a^T \alpha_b$) to ($\beta{'}_a^T\alpha_a < \beta{'}_a^T\alpha_b$), which break the alignment of the current model to old sample $a$. The superscript $T$ is a transpose operation that is often used for matrix multiplication.
When inter-modal deviation happens (Figure \ref{relation_dev}), 
the relative rotation of the representation space breaks the original modal alignment of the sample $a$, which makes the ($\beta_a^T\alpha_b > \beta_a^T\alpha_a$). 
From the perspective of the contrastive matrix, the element $M_{i,j}$ in the $i$,$j$ position of the contrastive matrix $M$ represents the similarity score of the $i$'th sample vision embedding and the $j$'th sample text embedding. Since the length of the representation vector is \textbf{1}, the similarity score $M_{i,j}$ also refers to the angle between the $i$'th sample vision embedding and the $j$'th sample text embedding. When the angle becomes larger due to spatial disorder, its similarity score within the model becomes smaller, which affects the multi-modal retrieval ability of the model. Because of this, the performance of CLIP$_{ct}$ drops significantly during continual training. Detailed mathematical derivations can be found in Appendix \ref{theoretical}. The value of the diagonal elements in the contrast matrix $M$ represents the angle between different modals of the same sample. The value of the off-diagonal elements represents the angle between the different modals of different samples in the CLIP's representation space. 
From an overall perspective, \textbf{the similarity distribution of the contrastive matrix $M$ is equivalent to the structure of the representation space of the model.} 

%% file: content/6_method.tex
\section{Alleviating Spatial Disorder}
\label{method}
\subsection{General continual CLIP training Setting}
Suppose we have used training dataset $D_0$ got a pre-trained model CLIP$_0$. And there is another vision-language dataset $D$. We split $D$ into $N$ sub-datasets $\{D_1, ..., D_N\}$, randomly and evenly, to simulate a stream of data and $D_t= \{(v_t^0,l_t^0),...,(v_t^{n},l_t^{n})\}$ denotes the training data in the training phase $t$, where $t \in \{1,2,...,N\}$. Then, we train the model CLIP$_0$ using this sub-datasets sequentially. The enocded $l_2$ normalized embeddings of vision and text is $V_t^i=E_V^t(v_t^i)$ and $L_t^i=E_L^t(l_t^i)$. When the model CLIP$_t$ is trained during the training phase $t$ using training data $D_t$, the previous sub-datasets $\{D_0, D_1, ..., D_{t-1}\}$ are no longer available. The joint training represents that training a CLIP$_{jt}$ from scratch using all data $D_{jt} = D_0 \cup D$.

\subsection{Mod-X: Maintain off-diagonal information-matrix}
To alleviate spatial disorder of the CLIP$_{ct}$ model during continual training. We propose a simple but effective new training framework: Maintain off-diagonal information-matrix (Mod-X). It boots the current CLIP model to retain the spatial alignment to past samples by distilling the contrastive matrix's off-diagonal information which is constructed by the model before and after continual training based on the current training data. The entire training framework is shown in Figure \ref{main_process}, where the $S^{i,j}$ means cosine similarity score of the $i$'th sample's vision embedding and the $j$'th sample's text embedding. The Contrastive module in Figure \ref{main_process} is a traditional InfoNCE loss \cite{baevski2020wav2vec} , which inherits from CLIP \cite{radford2021learning}. In the following, we mainly introduce our Spatial Alignment module.

\subsection{Spatial Alignment}
The diagonal elements in CLIP's contrastive matrix represent the similarity of the visual and language information of the current sample. The off-diagonal elements represent the similarity between the vision and language representation of the current sample and other samples. As mentioned in Section \ref{relationshape_sd_cm}, the distribution of the elements in the contrastive matrix represents the spatial distribution of representations between modalities of the model.
Therefore, we feel out the old model's representation space through the old model's contrastive matrix on the current training data. Then, selectively distill the old model's spatial distribution while training the current model. We construct contrastive matrix $M_{t-1}$ and $M_t$ using the last and current model CLIP$_{t-1}$ and CLIP$_{t}$ based on current sub-dataset $D_t$.
\begin{gather}
    M_{t-1}^{i,j} = CLIP_{t-1}(D_t) = s(E_V^{t-1}(v_{t}^i),E_L^{t-1}(l_{t}^j))  \\
    M_{t}^{i,j} = CLIP_{t}(D_t) = s(E_V^{t}(v_{t}^i),E_L^{t}(l_{t}^j)) 
\end{gather}

Where the $s(a,b) = a^Tb$ is the cosine similarity function. However, the last model's representation space for current data is not totally correct. For those misunderstood sample information (diagonal elements are not the largest in the current retrieval), we use the corresponding similarity information of the current model to replace them, thereby removing their influence during continual distillation. 
\begin{equation}
M_{t-1}^{(i,:)} = M_t^{(i,:)} ; \quad if \quad max(M^i_{t-1}) \neq i
\end{equation}
After that, we align the information matrix $M_{t-1}$ and $M_{t}$ using Kullback-Leibler Divergence \cite{csiszar1975divergence}.
\begin{equation}
L_{KL}^t(M_t,M_{t-1}) = -\sum M_{t-1} ln(\frac{M_t}{M_{t-1}})
\end{equation}

The final training loss can be written as $L_{Mod-X}$, and $\alpha$ is a hyper-parameter.
\begin{equation}
L_{Mod-X}^t = L_{InfoNCE}^t + \alpha L_{KL}^t
\end{equation}


%% file: content/7_experiment_change.tex
\section{Experiments}
\label{experiments}
\subsection{Datasets}
In the experiments, we use three different training datasets varying in scope and domain to evaluate the effectiveness of our Mod-X framework. \textbf{MS COCO Captions} \cite{lin2014microsoft}: MS COCO Captions (COCO) is a widely used image caption dataset. It contains 80K training images, 30K validation images, and 5K testing images (COCO(5K)). \textbf{Flickr30K} \cite{young2014image}: Flickr30K contains 30K training images and 1K test samples (Flickr30K(1K)) collected from Flickr, together with 5 reference sentences provided by human annotators. \textbf{ECommerce-T2I} \cite{M6-T} is a text-to-image e-commerce dataset that contains 90k training images and 5k testing images set (EC(5K)). Each image corresponds to a text description, and the description for each sample in the training and test sets does not repeat. Many detailed training settings and experiments (CC12M) can be viewed in Appendix \ref{Appendix_to_experiments}.

\begin{figure}[!h]
    \centering
	\subfigure{
	\centering
	\includegraphics[width=3.2in]{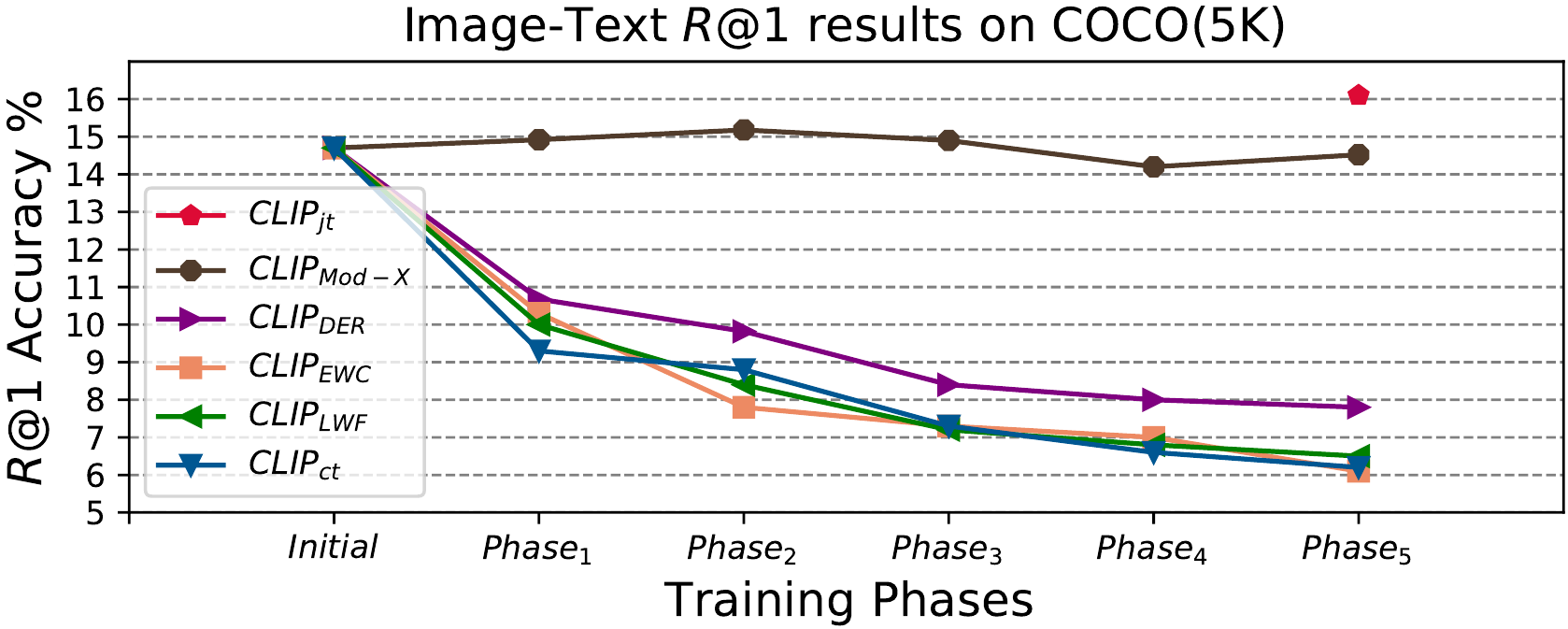}
	}
    
	\subfigure{
	\centering
	\includegraphics[width=3.2in]{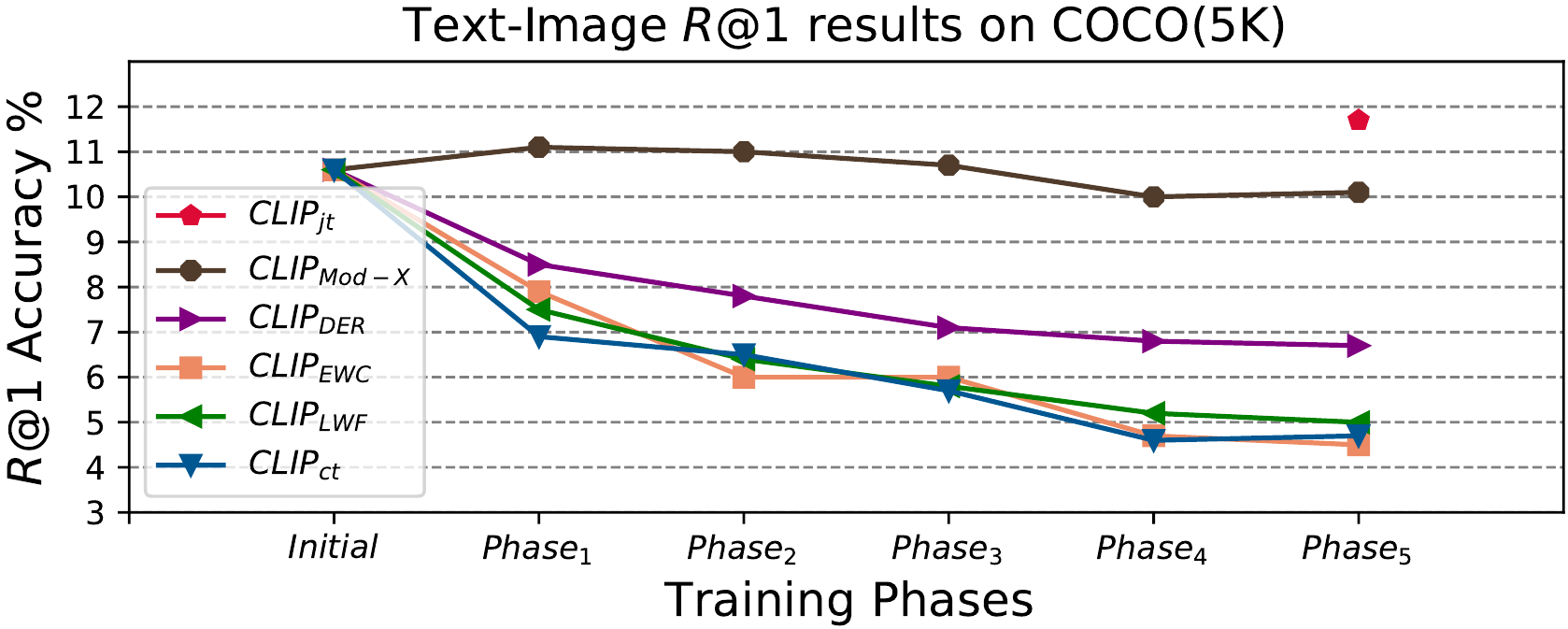}
	}
	
	\subfigure{
	\centering
	\includegraphics[width=3.2in]{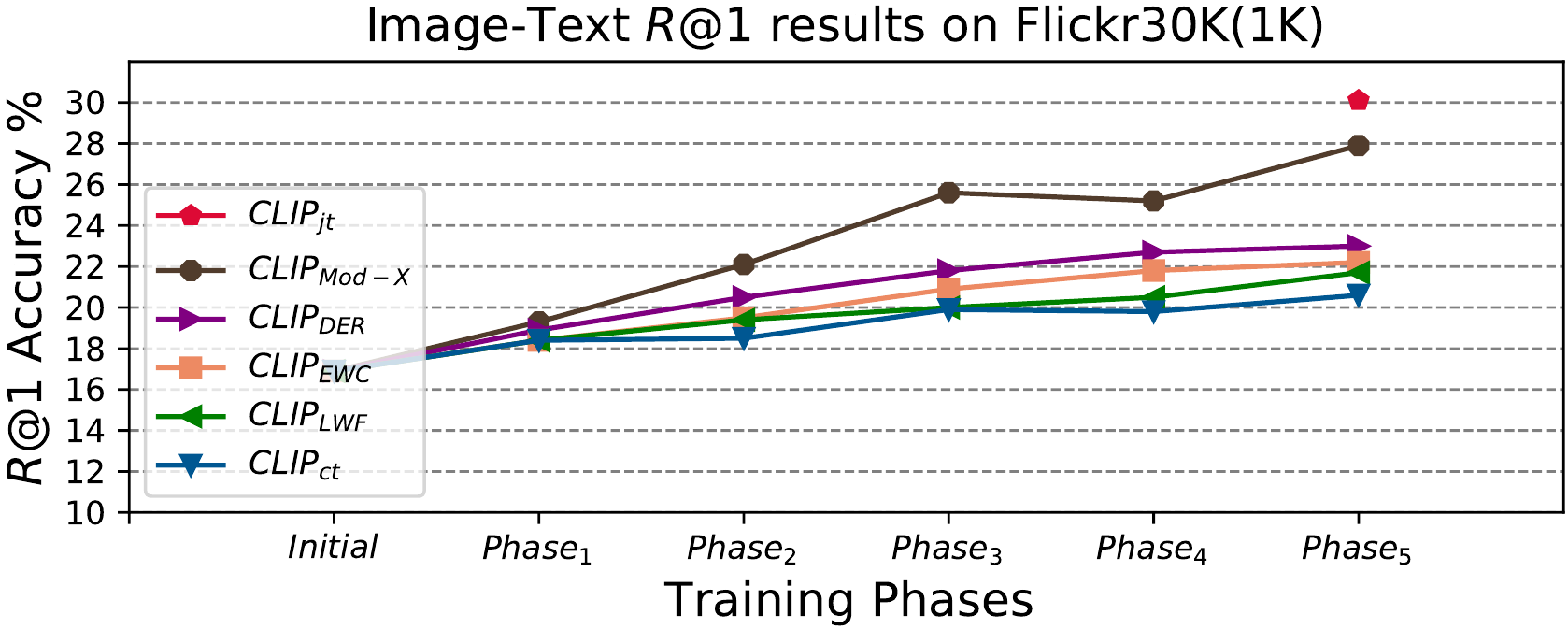}
	}
    
	\subfigure{
	\centering
	\includegraphics[width=3.2in]{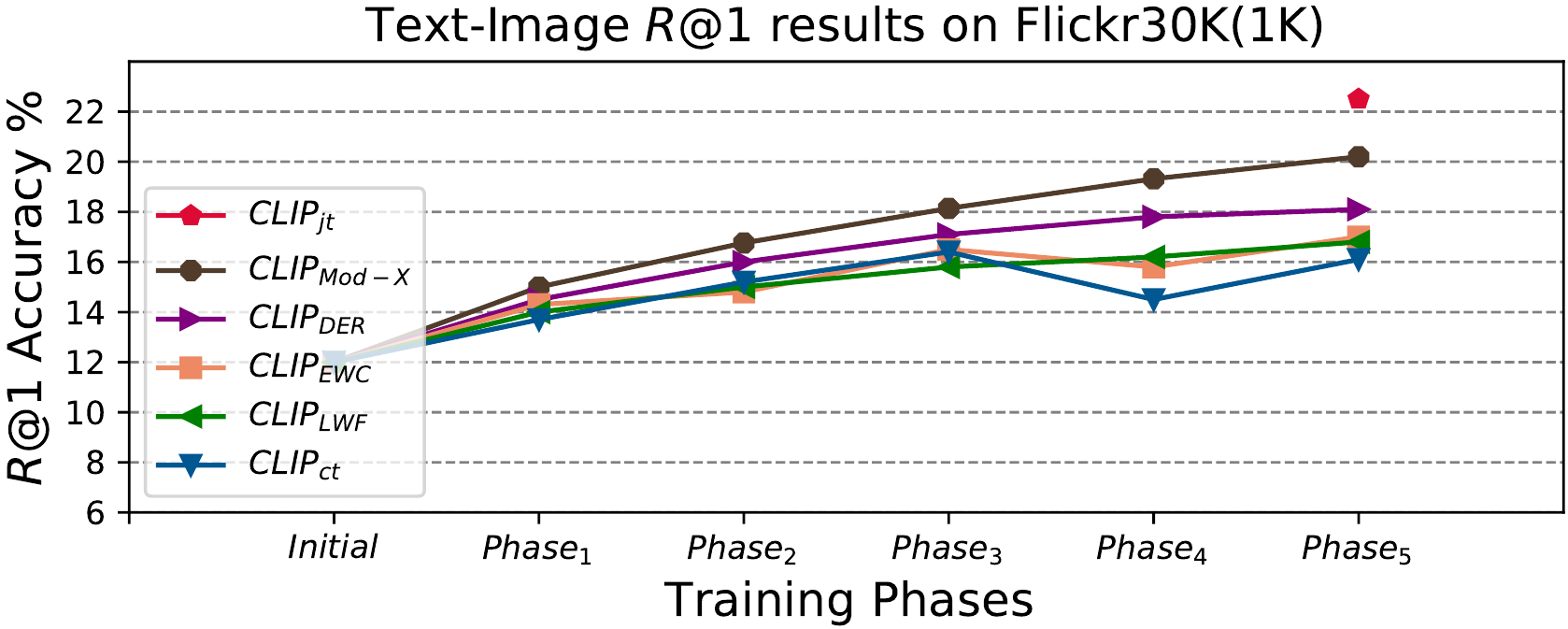}
	}
    \vspace{-2em}
	\caption{The multi-modal retrieval $R@1$ results of the different training strategies on COCO(5K) and Flickr30K(1K). The first two sub-figures show the retrieval $R@1$ performance on previous COCO dataset. The last two sub-figures show the $R@1$ results on current training domain Flickr30K.}
	\label{fig:modx_f30k_official}
\vspace{-1.em}
\end{figure}
\subsection{The performance in Exploratory Experiments}
\label{experiment_A}
\begin{figure*}[!t]
    \centering
	\subfigure{
	\centering
	\includegraphics[width=3.2in]{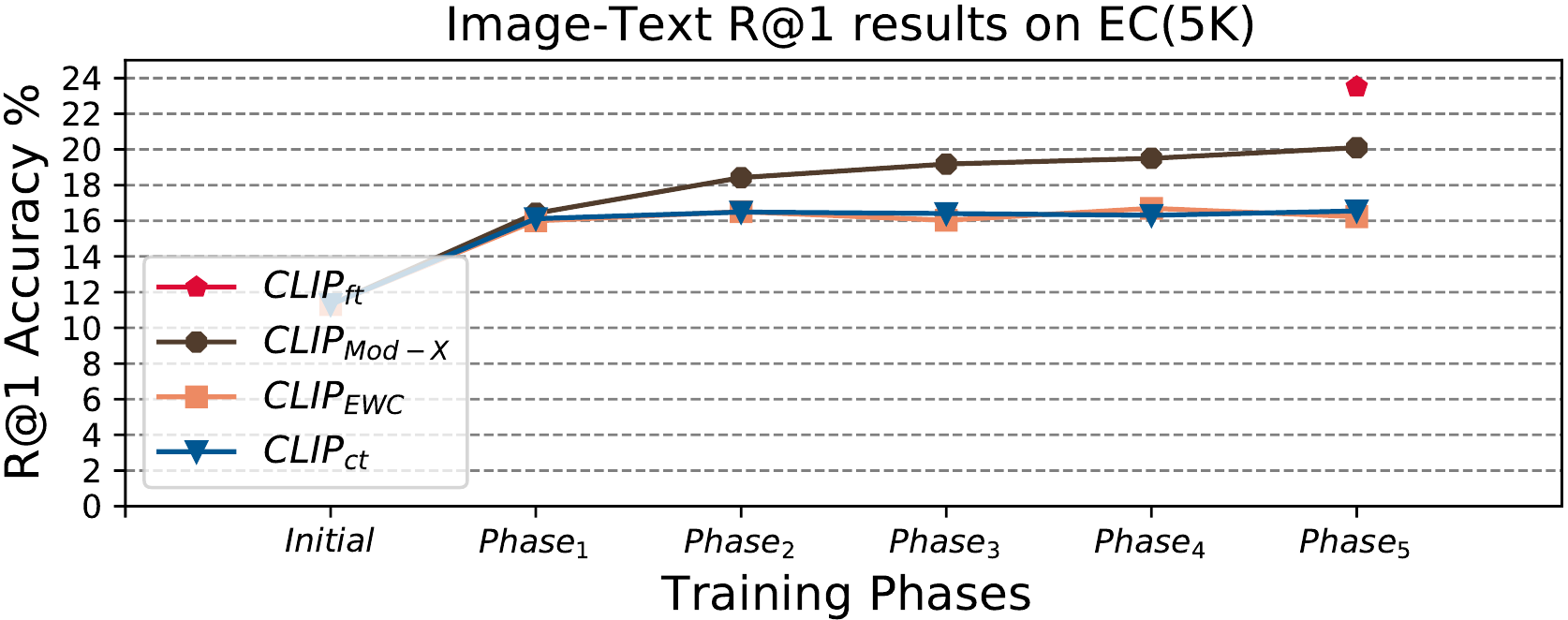}
	}
	\subfigure{
	\centering
	\includegraphics[width=3.2in]{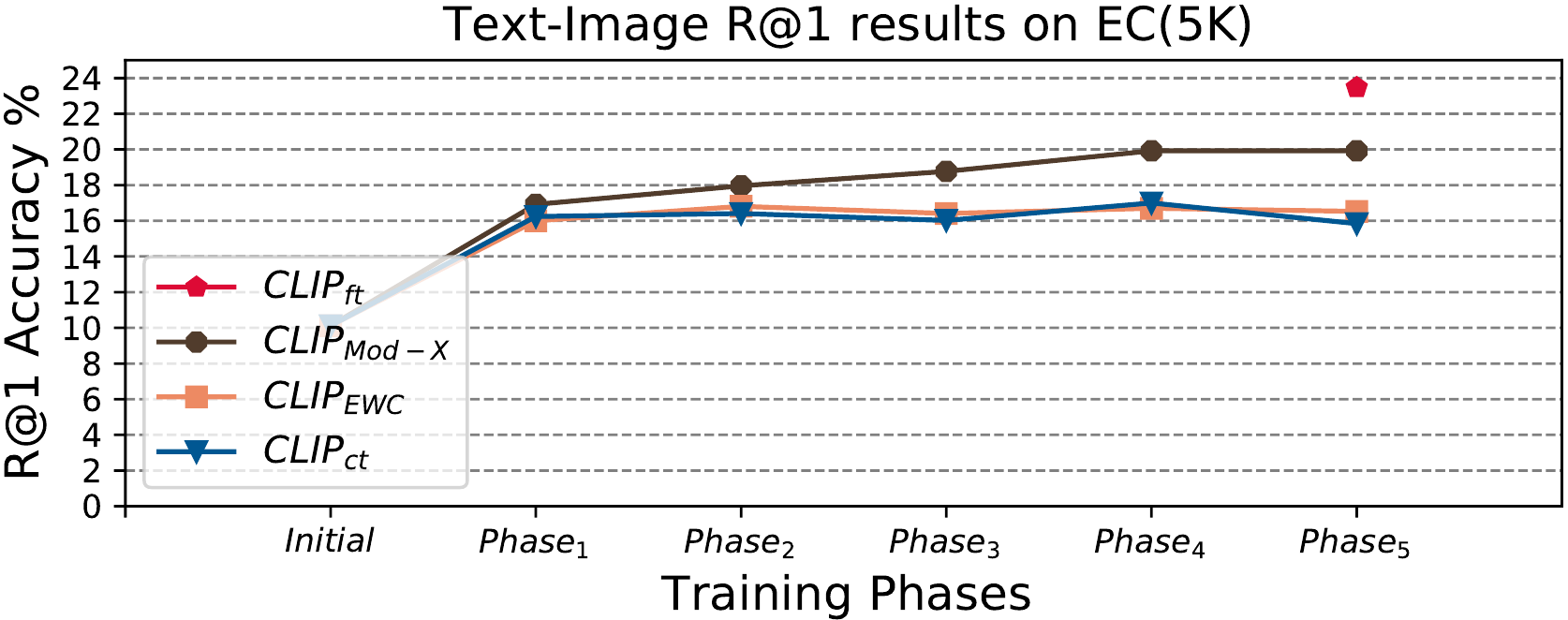}
	}
 
	\subfigure{
	\centering
	\includegraphics[width=3.2in]{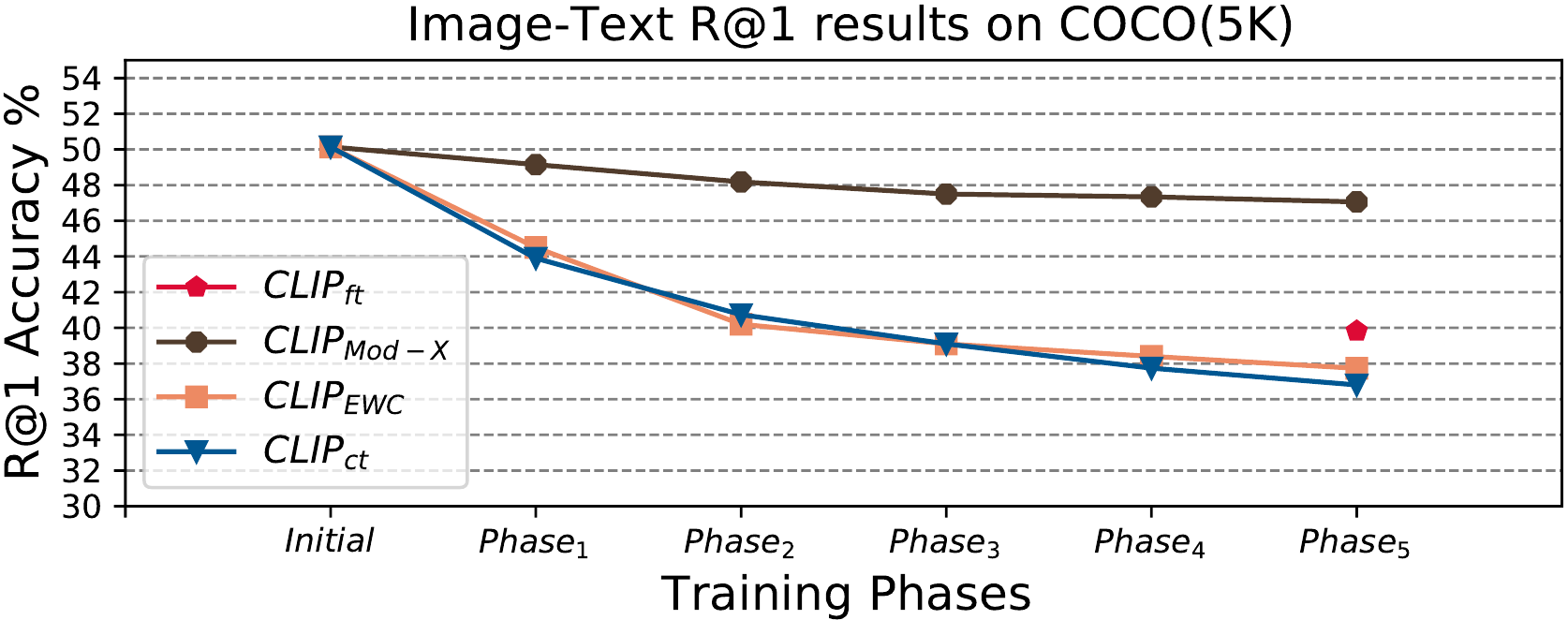}
	}
	\subfigure{
	\centering
	\includegraphics[width=3.2in]{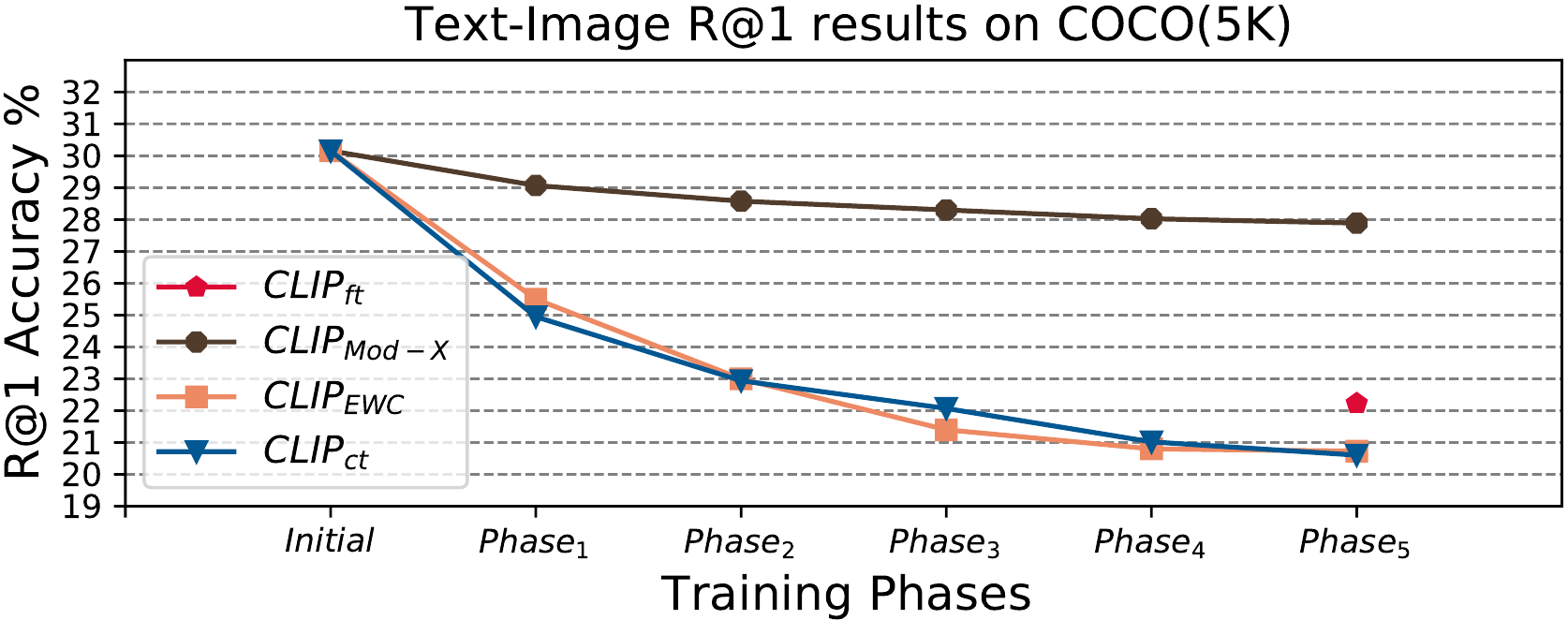}
	}
	
	\subfigure{
	\centering
	\includegraphics[width=3.2in]{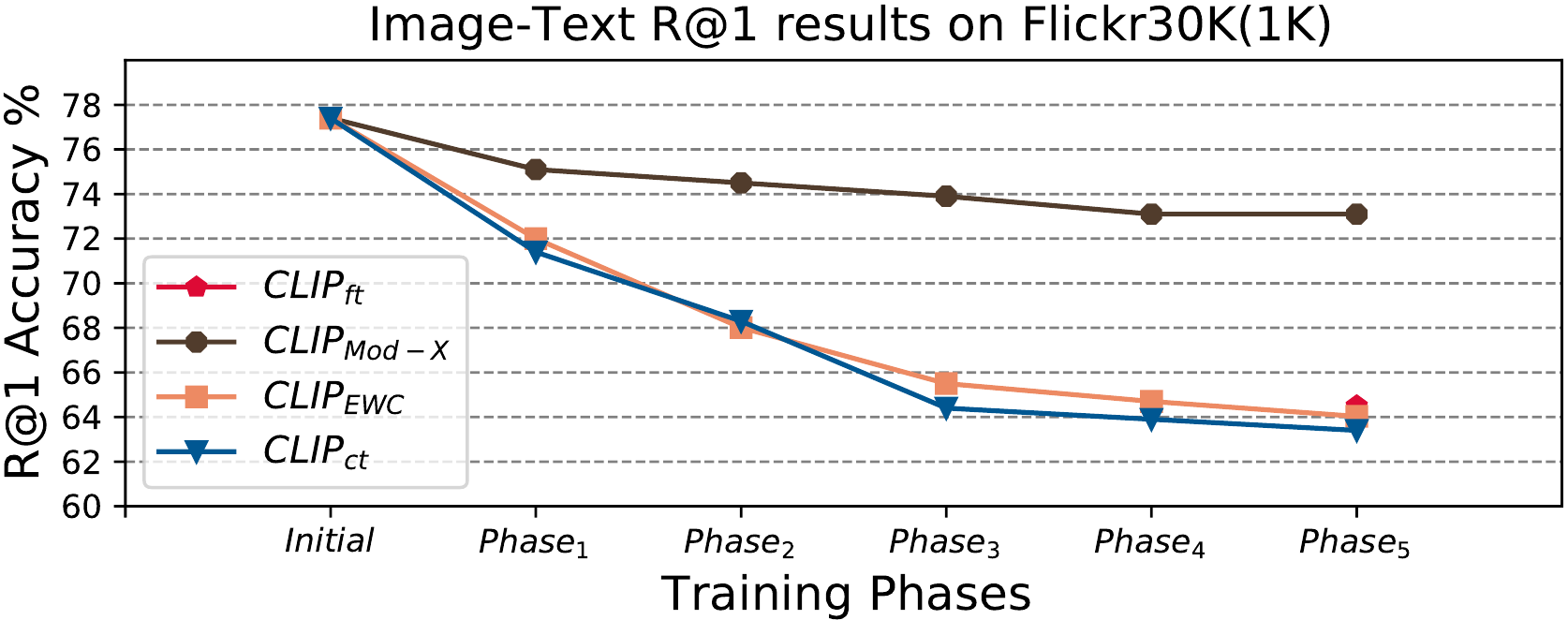}
	}
	\subfigure{
	\centering
	\includegraphics[width=3.2in]{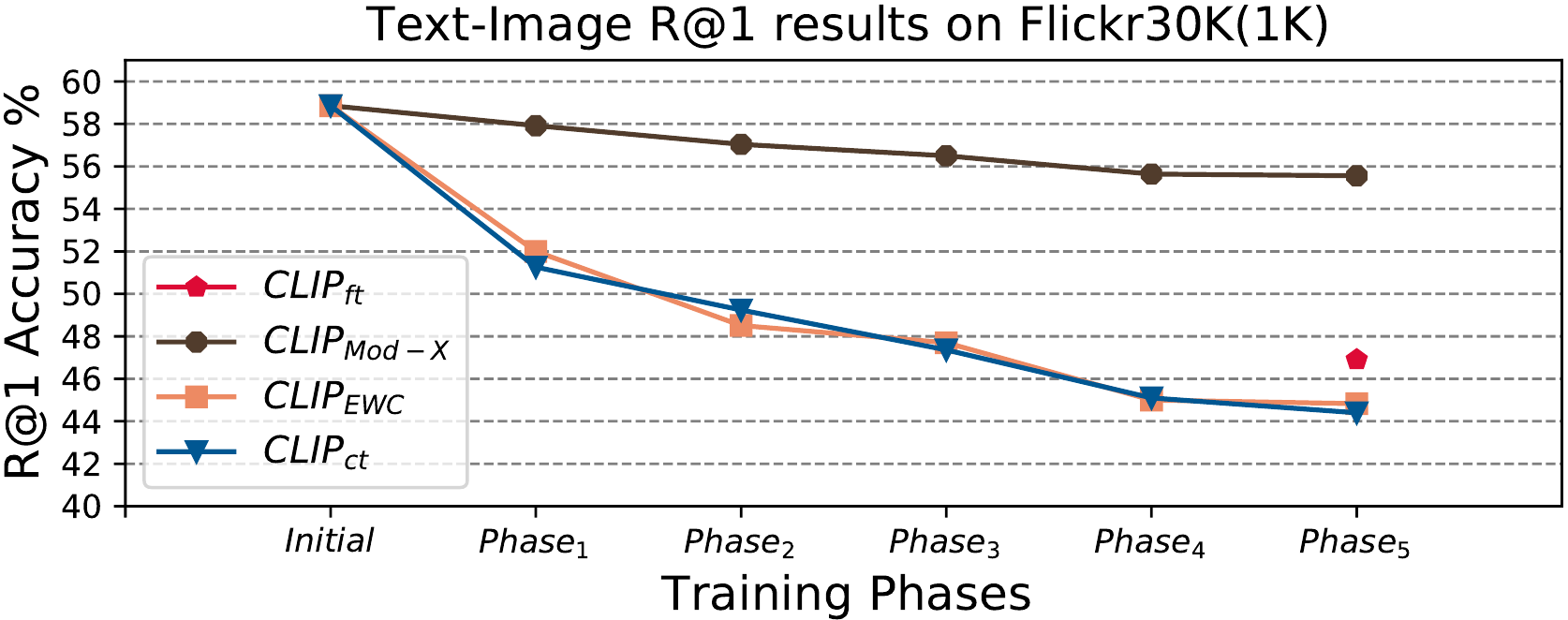}
	}
	\caption{The retrieval $R@1$ performance of different training strategies in each training phase on EC(5K), COCO(5K) and Flickr30K(1K). The CLIP$_{ft}$ is the fine-tuning results of CLIP$_{initial}$ on the full ECommerce-T2I dataset as an upper bound on CLIP$_{ct}$.}
	\label{fig:modx_ec_offifial}
    \vspace{-.5em}
\end{figure*}
Firstly, we follow the setup of the exploratory experiments described in Section \ref{explore_results}, comparing the results of our Mod-X framework with CLIP$_{ct}$, CLIP$_{EWC}$ and CLIP$_{jt}$ in the Flickr30K dataset. The CLIP$_{ct}$ means training CLIP continually without any other operation. The CLIP$_{jt}$ is training CLIP model in the joint dataset of COCO and Flickr30K, which is an upper bound for continual CLIP training. Since label information is not used in CLIP training, recent supervised continual training methods like iCaRL \cite{rebuffi2017icarl}, PodNet \cite{douillard2020podnet}, and Dyn \cite{yan2021dynamically} cannot be reproduced in such experimental settings. We compared our Mod-X with the typical continual learning strategy such as DER \cite{buzzega2020dark} (CLIP$_{DER}$), EWC \cite{kirkpatrick2017overcoming} (CLIP$_{EWC}$) and LWF \cite{li2017learning} (CLIP$_{LWF}$). Notably, to make LWF and DER work properly within the CLIP framework, we reproduced them using contrastive loss replaced their cross-entropy loss. The replay buffer size of DER is set to 3000. Figure \ref{fig:modx_f30k_official} shows the effect of our framework Mod-X (CLIP$_{Mod-X}$) and the performance of other training strategies at each training phase. At each training phase, the $R@1$ results of CLIP$_{Mod-X}$ on COCO(5K) did not show a significant drop, and the gap with the initial accuracy (\textit{Initial}) remained at $\pm 1\%$. Additionally, by comparing the retrieval performance of the CLIP$_{ct}$ and CLIP$_{Mod-X}$ on the current training data domain (Flickr30K), it can be found that the CLIP$_{Mod-X}$ is also significantly better than CLIP$_{ct}$ in continual fitting the current data domain. The low performance of CLIP$_{EWC}$ and CLIP$_{LWF}$ also shows that continual multi-modal training is more complex than single-modal supervised training. Due to the use of old training samples in memory buffer, the performance of DER is slightly better than LWF. However, its performance is still far from that of Mod-X. 
\begin{table}[!t]
\begin{minipage}[c]{1\linewidth}
\footnotesize
\setlength\tabcolsep{0.5pt}
\begin{tabular}{cc c}
\toprule
 Methods & COCO(I2T/T2I) & Flickr30K (I2T/T2I) \\
\midrule
$CLIP_{jt}$ &  16.1 / 11.7 & 	30.1 / 22.5 \\
$CLIP_{Mod-X}$ with 5000 &  15.3 / 11.0 & 29.1 / 21.7 \\
$CLIP_{Mod-X}$ with 3000 &  15.0 / 10.8 & 28.5 / 21.0 \\
$CLIP_{Mod-X}$ &  14.5 / 10.1	& 27.9 / 20.2 \\
\bottomrule
\end{tabular}
\end{minipage}
\vspace{-.5em}
\caption{
The table show the R@1 performance of the Mod-X framework in the final phase with a buffer size of 3000 or 5000. Since memory buffer strategy stores the model's knowledge from the input, it remains effective in the Mod-X framework which does not limit the input form.}
\label{memory_bank}
\vspace{-1em}
\end{table}
In Table \ref{memory_bank}, we presents the R@1 performance of the Mod-X framework in the final phase with a memory buffer size of 3000 or 5000. Before training on the Flickr30K, we randomly saved 1000 training samples from the COCO dataset into the memory buffer. Afterward, based on the size of the buffer, an equal number of current training samples were randomly selected and stored in the memory buffer after each continual training phase which is similar to previous works \cite{rebuffi2017icarl,buzzega2020dark,douillard2020podnet}. Since memory buffer strategy stores the model's knowledge from the input, it remains effective in the Mod-X framework which does not limit the input form. We can see from the results that as the number of old samples used increases, the performance of the Mod-X becomes closer to joint training. 

Beside of this, in order to show the performance of our Mod-X in high semantics correlations data sets, we adopt an approximate strategy to simulate class incremental setting in Flickr30K. Considering that the image labels are not available in Flickr30K, we used a pre-trained Imagenet1K model to automatically label the Flickr30K training data and divided it into 5 subsets, with each subset containing 200 classes. The R@1 results in the final phase have shown in Table \ref{cls_modx}, where the "cls" means "class incremental setting". From the results, it seems that continuous training with class incremental setting did not have a heavy impact on the effectiveness of the Mod-X. 
\begin{table}[!t]
\begin{minipage}[c]{1\linewidth}
\footnotesize
\setlength\tabcolsep{0.5pt}
\begin{tabular}{cc c}
\toprule
 Methods & COCO(I2T/T2I) & Flickr30K (I2T/T2I) \\
\midrule
$CLIP_{Mod-X}$ &  14.5 / 10.1	& 27.9 / 20.2 \\
$CLIP_{Mod-X}$ with cls &  13.8/9.8	& 27.4/19.7 \\
$CLIP_{ct}$ &  	6.2/4.7	& 20.6/16.1 \\
$CLIP_{ct}$ with cls &  6.3/4.7 & 18.1/15.7 \\
\bottomrule
\end{tabular}
\end{minipage}
\caption{
The table show the R@1 performance of the Mod-X framework in a class incremental setting. The results demonstrate that continuous training with class incremental setting did not have a heavy impact on the effectiveness of the Mod-X.}
\label{cls_modx}
\vspace{-1em}
\end{table}
In Appendix \ref{relationship_cm_ims_modx}, we show the spatial alignment in CLIP$_{Mod-X}$ and CLIP$_{ct}$, which demonstrates that the Mod-X framework can alleviate the spatial disorder well during continual CLIP training.

\subsection{The performance on special domain dataset ECommerce-T2I}
\label{Experiments_ECommerce-T2I}
To illustrate that the Mod-X framework is not only applicable to similar data domains, in this section, we compare the performance of different continual training frameworks on a specific e-commerce dataset ECommerce-T2I. We set the CLIP$_{vit32}$ with ViT-Base/32 vision encoder as the initial model, pre-trained using large-scale open-world datasets in \cite{OpenAi}. To simulate streaming data, we are dividing the entire ECommerce-T2I into five sub-datasets uniformly and randomly. Since the entire CLIP$_{vit32}$ pre-training dataset is not available, we use the fine-tuning results of CLIP$_{vit32}$ on the entire ECommerce-T2I dataset (CLIP$_{ft}$) as an upper bound on CLIP$_{ct}$.

The multi-modal retrieval $R@1$ results of CLIP$_{Mod-X}$, CLIP$_{ct}$, CLIP$_{EWC}$ and CLIP$_{ft}$ in each training phase are shown in Figure \ref{fig:modx_ec_offifial}. Comparing the $R@1$ performance of the CLIP$_{jt}$ and CLIP$_{ct}$ on the EC(5K) test set, it's clear that the training of the CLIP model is affected by the training data domain: The final $R@1$ results of CLIP$_{jt}$ on EC(5K) is 8.8\% (Image-Text) and 9.9\% points (Text-Image) higher than CLIP$_{ct}$. However, the retrieval results of CLIP$_{jt}$ on COCO(5K) and Flickr30K(1K) have dropped by more than 10\% points (comparing with CLIP$_{vit32}$ (\textit{Initial})) on average, which means that the performance of fine-tuning (one phase continual training) CLIP is also affected by the data domain. This is also verified by observations that the $R@1$ performance of CLIP$_{ct}$ performs lower than CLIP$_{jt}$. On the contrary, the CLIP$_{Mod-X}$ obtained after continual training by the Mod-X framework only has a tie drop of 3.3\% points in the $R@1$ retrieval results on COCO(5K) and Flickr30K(1K). What's more, the performance of the CLIP$_{Mod-X}$ on EC(5K) outperformed CLIP$_{ct}$ by 3.5\% (Image-Text R@1) and 4.2\% points (Text-Image R@1), respectively. The overall $R@1$ trend of CLIP$_{EWC}$ is similar to that of CLIP$_{ct}$ but is more unstable than CLIP$_{ct}$. All of this shows that Mod-X framework not only preserves the inter-modal spatial structure of old samples during the continual training but also improves the fitting ability of the CLIP in the current training data domain.

%% file: content/8_conclusion.tex
\section{Conclusion}
This paper discusses the feasibility of continuously training the CLIP model through streaming data. Then, by tracking the directional changes of the representation vectors in the continuously updated CLIP model, we explore and summarize these spatial variations as Spatial Disorder (SD), which can be divided into Intra-modal Rotation and Inter-modal Deviation. Moreover, we demonstrate how intra-modal rotation and inter-modal deviation lead to a performance decline for CLIP on cross-modal retrieval tasks in both empirically and theoretically. To alleviate the spatial disorder, we propose a simple yet effective continual learning framework Mod-X: \textbf{M}aintain \textbf{o}ff-\textbf{d}iagonal information-matri\textbf{X}. The experiments (in Section \ref{method}, \ref{experiments} and Appendix \ref{Appendix_to_experiments}) on commonly used datasets with different scales and scopes have illustrated the effectiveness of our method.